\documentclass{article}

\usepackage[preprint]{nips_2018} 

\usepackage[utf8]{inputenc} 
\usepackage[T1]{fontenc}    
\usepackage{hyperref}       
\usepackage{url}            
\usepackage{booktabs}       
\usepackage{amsfonts}       
\usepackage{nicefrac}       
\usepackage{microtype}      
\usepackage{xcolor}         

\usepackage{amsmath}

\usepackage{lineno}
\usepackage{lipsum}  
\usepackage{graphicx}

\usepackage[english]{babel}
\usepackage{algorithm}
\usepackage[noend]{algpseudocode}

\usepackage{bm}
\usepackage{wasysym}
\usepackage{amsmath}

\title{A toolkit for data-driven discovery of governing equations in high-noise regimes}

\author{Charles B. Delahunt, J. Nathan Kutz \\ 
Department of Applied Mathematics, University of Washington, Seattle, WA 98195-3925\\
\texttt\{delahunt, kutz\} @uw.edu}

\begin{document}

\maketitle

\begin{abstract}
We consider the data-driven discovery of governing equations from time-series data in the limit of high noise. 
The algorithms developed describe an extensive toolkit of methods for circumventing the deleterious effects of noise in the context of the {\em sparse identification of nonlinear dynamics} (SINDy) framework.  
We offer two primary contributions, both focused on noisy data acquired from a system  $\dot{\bm x} =  {\bm f} ({\bm x})$.

First, we propose, for use in high-noise settings, an extensive toolkit of critically enabling extensions for the SINDy regression method, to progressively cull functionals from an over-complete library and yield a set of sparse equations that regress to the derivate  $\bm{\dot{x}}$.  
This toolkit includes: 
(regression step) weight timepoints based on estimated noise, use ensembles to estimate coefficients, and regress using FFTs;
(culling step) leverage linear dependence of functionals, and restore and protect culled functionals based on Figures of Merit (FoMs).
In a novel Assessment step, we define FoMs that compare model predictions to the original time-series (i.e. ${\bm x}(t)$ rather than $\bm{\dot{x}}(t)$). 
These innovations can extract sparse governing equations and coefficients from high-noise time-series data (e.g. 300\% added noise). 
For example, it discovers the correct sparse libraries in the Lorenz system, with median coefficient estimate errors equal to  1\%$-$3\% (for 50\% noise), 6\%$-$8\% (for 100\% noise);  
and 23\%$-$25\% (for 300\% noise).
The enabling modules in the toolkit are combined into a single method, but the individual modules can be tactically applied in other equation discovery methods (SINDy or not) to improve results on high-noise data. 

Second, we propose a technique, applicable to any model discovery method based on $\dot{\bm x} =  {\bm f} ({\bm x})$, to assess the accuracy of a discovered model in the context of non-unique solutions due to noisy data.
Currently, this non-uniqueness can obscure a discovered model's accuracy and thus  a discovery method's   effectiveness.
We describe a technique that uses linear dependencies among functionals to transform a discovered model into an equivalent form that is closest to the {\em true} model, enabling more accurate assessment of a discovered model's accuracy. 
\end{abstract}

%


\section{Introduction}

\footnotetext{This work has been submitted to the IEEE for possible publication. Copyright may be transferred without notice, after which this version may no longer be accessible.} 

The derivation of governing equations for physical systems has dominated the physical and engineering sciences for centuries.  Indeed, it is the dominant paradigm for the modeling and characterization of physical processes, engendering rapid and diverse technological developments in every application area of the sciences.  Since the mid 20th century, governing equations have become even more influential due to the rise of computers and scientific computing.  Scientific computing allows one to emulate diverse and complex systems that are high-dimensional, multi-scale and potentially stochastic in nature.  In modern times, the rapid evolution of sensor technologies and data-acquisition software/hardware, broadly defined, has opened new fields of exploration where governing equations are difficult to generate and/or produce.  Biology and neuroscience, for instance, easily come to mind as application areas where first-principle derivations are difficult to achieve, yet data is now becoming abundant and of exceptional quality.   Measured  coarse-grained macroscopic behavior is also often difficult to derive or characterize from known microscopic descriptions.  The ability to discover governing equations directly from time-series data is thus of paramount importance in many modern scientific and engineering settings.  Confounding the discovery process is the ubiquity of noisy experimental data. The scope of this work is centered around the data-driven discovery of a system's governing equations given highly noisy experimental data.

Time-series measurements for which the signal-to-noise ratio is low represents significant challenges for any analysis of the underlying signal.  While the techniques described here can apply to various discovery or signal processing methods, we consider in particular the {\em sparse identification of nonlinear dynamics} (SINDy) algorithm~\cite{Brunton2016pnas,bruntonKutzBook2019}.  SINDy is a regression method that leverages time-series data to discover the governing equations of a system of differential equations (or partial differential equations~\cite{Rudy2017sciadv,Schaeffer2017prsa}
\begin{equation}
\label{eqnGeneralSindy}
 \dot{\bm{x}} = \bm{f(x)} ,
\end{equation}
with state $\bm{x} \in R^n$ and smooth dynamics $\bm{f(x)}: R^n \rightarrow R^n$.
SINDy assumes that the governing equation of each variable $x_j$ is a linear combination of a small number of terms, i.e. 
\begin{equation}
\label{eqnSubscriptSindy}
\dot{x}_j = \sum_{k \in S_j}  \xi_{k} f_k(\bm{x}) 
\end{equation}
where $f_k(\bm{x})$ are candidate library terms of the dynamics and the $\xi_k$, representing their weights or loadings, are assumed to be mostly zero.  Indeed, only a few non-zero terms $\xi_k$ are assumed to be relevant in discovering  the parsimonious dynamical model (Eqn \ref{eqnGeneralSindy}).   The basic SINDy method starts with a pre-defined, overcomplete library of functionals $F = \{f_i(\bm{x})\}$ (e.g. typically polynomials up to some degree), and assumes $\dot{\bm{x}}(t) = \Xi F(\bm{x}(t))$ for a coefficient matrix $\Xi$.
It then culls coefficients until $\Xi$ is sparse.  Then for a given variable $x_j$ the non-zero entries in the $j^{th}$ row of $\Xi$ correspond to the $\{\xi_{k}\}$ in (Eqn \ref{eqnSubscriptSindy}).
The original formulation of SINDy estimated and culled functionals using {\em Sequentially Thresholded Least Squares} (STLSQ) \cite{Brunton2016pnas,zhang2019convergence}, as a tractable alternative to $\ell_1$ regression.
Other culling methods have been developed. In particular \cite{champion2020unified} used a robust method termed SR3 (sparse relaxed regularized regression) to extract the sparse, non-zero coefficients of the dynamical model.   

Since its introduction, SINDy has been applied to a wide range of systems, including for reduced-order models of fluid dynamics~\cite{Loiseau2017jfm,Loiseau2018jfm,loiseau2020data,guan2020sparse,deng2021galerkin,callaham2021empirical,callaham2021role} and plasma dynamics~\cite{Dam2017pf,kaptanoglu2020physics}, turbulence closures~\cite{beetham2020formulating,beetham2021sparse,schmelzer2020discovery}, nonlinear optics~\cite{Sorokina2016oe}, numerical integration schemes~\cite{Thaler2019jcp}, discrepancy modeling~\cite{Kaheman2019cdc,de2019discovery}, boundary value problems~\cite{shea2021sindy}, multiscale dynamics~\cite{champion2019discovery}, identifying dynamics on Poincare maps~\cite{bramburger2020poincare,bramburger2021data}, tensor formulations~\cite{Gelss2019mindy}, and systems with stochastic dynamics~\cite{boninsegna2018sparse,callaham2021nonlinear}.  It can also be used to jointly discovery coordinates and dynamics simultaneously~\cite{Champion2019pnas,kalia2021learning}.
The integral formulation of SINDy~\cite{Schaeffer2017pre} has also proven to be powerful, enabling the identification of governing equations in a weak form that averages over control volumes; this approach has recently been used to discover a hierarchy of fluid and plasma models~\cite{Reinbold2020pre,gurevich2019robust,alves2020data,reinbold2021robust}. 
The open source software package, PySINDy\footnote{\url{https://github.com/dynamicslab/pysindy}}, has been developed in Python to integrate the various extensions of SINDy~\cite{deSilva2020JOSS}.  For actuated systems, SINDy has been generalized to include inputs and control~\cite{Brunton2016nolcos}, and these models are highly effective for model predictive control~\cite{Kaiser2017arxivB}.  
It is also possible to extend the SINDy algorithm to identify dynamics with rational function nonlinearities~\cite{Mangan2016ieee,kaheman2020sindy}, integral terms~\cite{Schaeffer2017pre}, and based on highly corrupt and incomplete data~\cite{Tran2016arxiv,champion2020unified}.  
SINDy was also recently extended to incorporate information criteria for objective model selection~\cite{Mangan2017prsa}, and to identify models with hidden variables using delay coordinates~\cite{Brunton2017natcomm}.  The diversity of methods and applications highlight the broad reach and flexibility of the underlying regression architecture (Eqn \ref{eqnSubscriptSindy}).  The high-noise methods introduced here can help with all these formulations since noise severely limits the usefulness of SINDy in such a parameter regime. \\ \\


\subsection{Challenges of SINDy}
\label{sectionDrawbacks}
This paper seeks to address common challenge points for the SINDy method and its variants. These include:

\begin{enumerate}
\item SINDy's regressions fit the derivatives $\bm{\dot{x}}$, not the original time-series $\bm{x}$. 
Given noisy data, this has (at least) two effects: 
First, the solutions are not unique (as noted in \cite{Schaeffer2017pre}), especially given an overcomplete starting library.
There can exist several plausible sparse libraries, and for a fixed sparse library a range of coefficients, each of which fit the derivatives  well but give different estimates $\bm{\hat{x}}(t)$ when evolved in time.  
Thus, fitting the derivatives is not a sufficient method to find a sparse model which reproduces the original time-series. 
\item What level of sparsity to enforce is not known, so the output of the SINDy algorithm is often too dense or too sparse and hyper-parameter tuning is required. 
A standard solution is to do multiple regressions, sweeping the sparsity parameter $\lambda$, then choose between the resulting models by some method, e.g. (i) expert assessment of models on a Pareto front \cite{Mangan2016ieee}; (ii) finding the knee where error (vs $x(t)$) stabilizes \cite{Schaeffer2017pre}; or (iii) examining error of the fit to $\dot{x}$ via cross-validation \cite{boninsegna2018sparse} or on a holdout trajectory \cite{kaheman2020sindy}.
  \item The progressive culling of functionals from the library is greedy, so if a vital functional is culled early the method cannot recover (observed  for SINDy by \cite{boninsegna2018sparse}, and for Lasso by \cite{su}).
\item Library functionals are culled based on their coefficients' magnitudes. 
This penalizes functionals with large-valued $f_i(t)$.
For example, if a time-series $x(t)$ hovers around the value 10, then the functional $x^2(t)$ can have a coefficient $\xi_2$ that is 10$\times$ smaller than the coefficient $\xi_1$ of $x(t)$ while exerting the same effect in the regression, since $\xi_2x^2(t) \approx \xi_1 x(t)$.
Since functionals with small coefficients are culled first, $x^2(t)$
 will be culled despite having equal impact in Eqn \ref{eqnSubscriptSindy}.
A Gaussian prior can be imposed on the coefficients \cite{hirsh, zhang}, but this may be a poor match for natural systems, which can have large variation in coefficient magnitudes.
Library functional time-series $f(t)$ can be normalized to unit variance \cite{quade}, but this risks imposing an opposite bias on coefficient size, where $x^2(t)$ has equal footing with $x(t)$, even though its actual coefficient would be 10$\times$ smaller, and in noisy conditions more volatile.
\item Noisy data is  especially problematic because both the regression target $\bm{\dot{x}}$ and the library functionals' time-series $f_i(t)$ are estimated from this data.
Existing approaches to handling noise are discussed in the next section.
\end{enumerate}


\subsection{Dealing with high noise in the SINDy context}
High noise is the central challenge addressed in this paper. 
Noisy data disrupt SINDy primarily by compromising derivative estimates and by distorting estimates of library functional values $f_i(\bm{x}(t))$.
The original SINDy formulation \cite{Brunton2016pnas} handled 12\% Gaussian noise   ($\sigma_{noise} / \sigma_{data} = 0.12$) in the Lorenz system 
using total variation regularization of derivative estimates, with some error in the final estimated coefficients $\Xi$.
The special case of partial corruption (i.e. some timepoints contain noise, while others are noise-free) has also been considered~\cite{champion2020unified,Tran2016arxiv}. 
By trimming outliers, Champion et al~\cite{champion2020unified} handled 10\% of datapoints corrupted with heavy noise (spreading to 30\% of derivative values corrupted)  in the Lorenz system. 
The theoretical bounds of the least median of squares method \cite{rousseeuw}) suggests that higher levels of corruption might be addressable with this approach.
Tran et al~\cite{Tran2016arxiv} recovered correct equations and coefficients for the Lorenz system given up to 72\% of points corrupted, subject to (i) low levels of noise; (ii) sufficiently long time-series; and (iii) a corruption pattern alternating long clean segments with corrupted segments.   
Rudy et al~\cite{rudySmoothing2019} assumes exact knowledge of the governing sparse library but not its coefficients, and then includes a loss term measuring whether the current model satisfies the implied dynamics. 
This method identified correct coefficients for the Lorenz system given roughly 120\% noise (including a non-zero mean case), suggesting that it might serve as an effective second stage in a chain where the first stage  identifies the correct sparse library but not the correct coefficients.

Noise in PDEs is especially challenging for SINDy, because noise amplifies with each (spatial) derivative calculation \cite{Rudy2017sciadv, Reinbold2020pre}.
\cite{Rudy2017sciadv} found that even low noise resulted in much decreased coefficient estimates (e.g. magnitude roughly halved) of discovered sparse library functionals. 
\cite{ahnert} notes that a simple moving average of noisy data always reduces estimates of extrema, which would reduce the functional coefficients $\xi_k$ in Eqn \ref{eqnSubscriptSindy}.
\cite{Schaeffer2017pre} addresses noise via integration, so that the sparse regression involves not the noisy $\dot{x}$ and $f(t)$s but rather their much cleaner integrated versions. 
This weak formulation has potential value for PDEs, where noise amplifies with each partial derivative but can be effectively mitigated by integration.
It handled 3\% noise in the Lorenz system (this result may understate the value of the method for PDEs).
\cite{Reinbold2020pre} also applies integration, via multiplication by a smooth kernel along with integration by parts, to reduce effects of noise. 
This method handled 5\% noise in a reaction-diffusion equation, and usually recovered the correct sparse library elements (with coefficient errors) even at 10\% noise.


\begin{figure}[t]
\centering 
\centerline{
\fbox {
\includegraphics[width=1\linewidth]{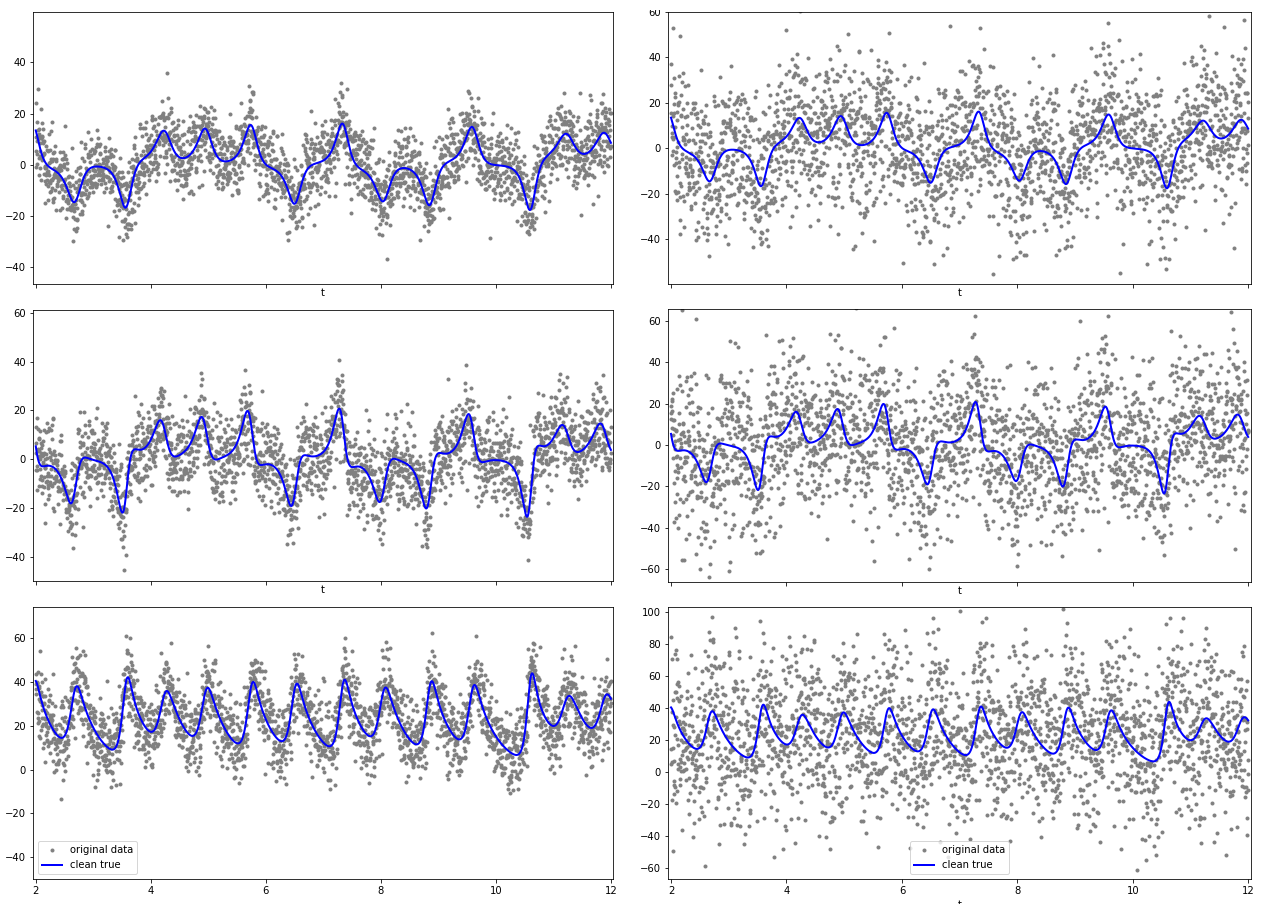}  
}
}
\caption{\small { \textbf{The noisy data regime.} 
Lorenz time-series ($x, y, z$) with added white noise. 
Blue lines are true, clean trajectories. 
The effective added noise levels, $\sigma_{noise} / \sigma_{data}$, are $\approx$100\% (left) and $\approx$300 \% (right).
The toolkit described here discovers the correct functionals for each variable, with median coefficient estimate errors equal to 6\% to 8\%, and 6\% (for 100\% noise); and 23\% to 25\% (for 300\% noise).
See Results for details.
} }
\label{figIncreasingNoiseLorenz}
\end{figure} 


\subsection{A toolkit for noisy data}

This paper considers the case of added white noise equivalent to 50\% to 300\% gaussian noise, affecting all data points.
See Figure \ref{figIncreasingNoiseLorenz} for  examples.
We consider various systems described by polynomial libraries, including Lorenz, harmonic oscillators, and the Hopf Normal form. 
PDE or rational function systems are not addressed here.
%
We apply an engineering lens to SINDy to address the exigencies of noisy data, and describe a toolkit of novel, practically-based techniques, including:

In the Regression step we weight timepoints based on estimated noise, use ensembles to estimate coefficients, and regress using FFTs.
In the Culling step 
we rescale coefficients, leverage linear dependence of functionals, and restore and protect culled functionals based on Figures of Merit (FoMs).
In a novel Assessment step we define FoMs that compare model predictions to the original time-series (i.e. to ${\bm x}(t)$ rather than $\bm{\dot{x}}(t)$).

We emphasize that the individual techniques can operate separately, and are intended to be incorporated tactically into other  frameworks, including but not restricted to SINDy. 
Here we present the toolkit combined into a single architecture, and it can be used as such; but it is really several independent modules strung together to produce an effective overall architecture for model discovery in the high noise limit.
As with SINDy, we wish to discover a sparse set of governing equations by fitting functionals to the derivatives of the system.
We start with an overcomplete library of functionals, and progressively cull  functionals with small magnitude coefficients via STLSQ, until we achieve a sparse library of functionals that accurately model the system dynamics. 
The several differences from existing SINDy programs are described in our methods. 
 

\subsection{Contributions of this paper}
We offer two main contributions, both applicable to discovery methods generally, not just to SINDy.
First, we describe a toolkit of techniques that enable accurate discovery of sparse governing equations in very high-noise settings (50 - 300\% added noise).
The various modules and ideas in the toolkit can be  separately  inserted as needed into other  discovery frameworks to improve their performance in high-noise regimes.
Second, we address the problem of non-uniqueness of solutions found from high-noise data.
A discovered model can appear to have incorrect functionals and/or coefficients, while it is in fact equivalent (transforming by linear dependencies) to a form that is close to the ``true'' model \cite{champion2019data}.
However, no method currently exists (to our knowledge) to do this.
We propose an automated technique to linearly transform a discovered model into an equivalent form, subject to the constraints of the input data, that is a closest match to the ``true'' model.
This enables better assessment of a method's actual effectiveness at discovering governing equations.



\subsection{Evaluation of results}
Clear, universal metrics of success  are perhaps not possible in the context of discovery of governing equations from noisy data.
First,  solutions in noisy settings are non-unique (cf section \ref{sectionDrawbacks}), and require special assessment methods, which we introduce and describe below.
Second, different use-cases have diverse needs.
At least three desiderata, of increasing difficulty, are used in the literature:

\begin{enumerate}
\item We wish to  identify the correct sparse library, i.e. the non-zero functionals in the governing equations. 
Noise complicates this task because linear dependencies between the functionals in an over-complete library, i.e. $f_k(t) \approx \sum_{i \neq k} \beta_i f_i(t) \text{ for most } t$, mean that multiple sets of functionals can represent the system within a noise-induced margin of error.
\item We wish to accurately estimate the coefficients of the  functionals.
The presence of noise complicates this in various ways:
(i) Regressions on different subsets of points will yield different coefficients, even for the same set of functionals, as noise distorts the trajectory $\bm{x}(t)$and the derivatives $\bm{\dot{x}}(t)$ being fitted.
(ii) Any smoothing used to de-noise the data tends to add artifacts and also reduce the sharper transitions (e.g.remove peaks in time-series) which reduces derivatives' apparent magnitudes. 
\item We wish to accurately predict trajectories in different parts of the state space. 
In non-linear or chaotic systems, small differences in model coefficients may yield predicted trajectories $\bm{\dot{x}}(t)$ which diverge from clean ground-truth even while they live on the same attractor and exhibit the same qualitative behaviors (e.g. attractors, cycles, magnitudes, dominant frequencies, state-space histograms).
\end{enumerate}

We emphasize that solutions are not unique for noisy systems, due to over-complete libraries and quasi-linear dependence of functionals within the noise envelope. 
Thus, various ``true'' functionals can be absent from the discovered model but still lie within the linear span of the discovered (perhaps ``incorrect'') functionals.
Also, the ``true'' functionals may themselves have linear dependencies, so that their coefficients can vary noticeably with only slight effect on trajectory behavior. 
A discovered solution can thus appear quite wrong, both in terms of (1) and (2) above, while  in fact being a linear transform of a solution close to the ``true'' system, and/or also producing correct trajectories (item 3).
The technique  in section \ref{sectionLinearSpanAssessment} finds an equivalent model within the linear span of the discovered model that is closest to the ``true'' solution, which improves the accuracy of (1) and (2) above.

We also note that different use-cases require different definitions of success.
For a domain expert seeking to use experimental data to gain insight into a system under study, items (1) and partially (2) might suffice. 
To simulate systems for experimental study, items (2) and partially (3) matter.
For prediction and control, item (3) is important.
In this paper we report quantitative results for (1) and (2), and qualitative results for (3). 
Re item (3), we found that in high-noise chaotic systems the predicted trajectories of discovered models, even with accurate sparse libraries and coefficients, tended to fall off of (and onto) the true trajectories, while maintaining very similar qualitative behavior .

\section{Methods}

This section describes the  high-noise toolkit, as follows: 
We first briefly walk through how the framework processes a dataset, listing the several modules.
We then describe in detail the individual modules, grouped according to context. 
Miscellaneous other modules are described in the Appendix, because they either (i) gave benefit but require an existing SINDy method (e.g. the pySINDy Python package \cite{deSilva2020JOSS}); or (ii) showed no clear benefit.

A full Python codebase for the toolkit  can be found at \cite{codebaseForHighNoiseSindyToolkit}. 
The toolkit is also in process of being added to the pySINDy package.
 
\subsection{Toolkit walk-through}

The toolkit has three main stages, each with multiple steps: (1) Preparation; (2) Iterations; (3) Final tasks. 
In the list below, standard SINDy steps are marked with $(**)$, other steps reference the relevant subsection. 
Given time-series $\{x_j(t)\}$, the procedure works as follows: \\

\noindent \textbf{1. Preparation:}
Done for each variable $x_j$ separately.
\begin{enumerate}
\item  \textbf{Smooth} each $x_j$ with a Hamming filter (or other method) \ref{sectionSmoothing} 
\item  Define a library of functionals $\{f_i\}$, possibly different for each $x_j$ $(**)$
\item Calculate each $\dot{x}_j$ (e.g.via Runge-Kutta) and each $f_i$, using the smoothed $x_j$s $(**)$
\item Calculate base weights for the timepoints (for regressions) \ref{sectionWeightTimepoints}
\item  Calculate {\em Fast Fourier Transform} (FFT) of each $\dot{x_j}$ (for regressions) \ref{sectionFftRegression}
\item  Calculate median values of each $f_i$ (for rescaling coefficients during culling) \ref{sectionWeightCoefficients}
\item Calculate histograms and FFT power spectra for each $x_j$ (for Figures of Merit).
\end{enumerate}

\noindent \textbf{2. Iterations:}\\
Each iteration includes three key stages: (i) Regression, (ii) Figures of Merit, and (iii) Culling.
Some steps of Regression are done for each variable $x_j$ separately.
Figures of Merit, Culling, and some steps of Regression are done on the full model (all variables together).
Below, ``active functionals'' correspond to $\{i,j\}$ pairs where $\xi_{ij} \neq 0$ (so if $\xi_{i1}, \xi_{i2} \neq 0$, then $f_i$ for $x_1$ and $f_i$  for $x_2$ are distinct active functionals).\\

\noindent \textbf{(i) Regression:} 
\begin{enumerate}
\item  \textbf{Choose \textit{n} subsets of timepoints for regression} (e.g. $ n = 17$) \ref{sectionBoosting}.  
For each subset $S$:
\item For each $x_j$:  \textbf{Weight the timepoints during regressions.} Use the timepoint weights based on estimated noise during regressions  \ref{sectionWeightTimepoints}
\item For each $x_j$:  \textbf{Remove timepoints with too-large ratios} of  active $f_i$ values from each $S$, since these can destablize the regression onto $\dot{x}_j$  \ref{sectionIgnoreExtremeFunctionalRatioPoints}
\item For each $x_j$: 
\textbf{Regress onto FFTs.} Use  a target consisting of both $\dot{x_j}(t)$ and the FFT values for $\dot{x_j}(t)$ for linear regression to find coefficients $\xi_{ij}$.
Regression onto $x_j(t)$ is standard $(**)$;  regression onto FFTs  is not \ref{sectionFftRegression}
\item For each $x_j$: 
\textbf{Combine the $\xi_{ij}$ estimates.} For each $f_i$, there are $n$ estimates for its coefficient $\xi_{ij}$, one from each subset $S$. 
Set the estimate of $\xi_{ij}$ equal to their median  \ref{sectionBoosting}
\end{enumerate}

\noindent \textbf{(ii) Figures-of-Merit (FoMs):} 
We calculate FoMs at each iteration, to help assess which model(s) in the progressively-sparser sequence yield the best estimated time-series $\hat{x}_j$ (distinct from the fit to derivatives $\dot{x}_j$).
\begin{enumerate} 
\setcounter{enumi}{5}
\item \textbf{Evolve the current  model} $ \{ \hat{\dot{x}}_j = \sum_{i} \xi_{ij} f_{i}, \forall j \}$ over train and validation trajectories \ref{sectionEvolve}.
\item \textbf{Calculate  FoMs}, based on the evolutions \ref{sectionFoms}
\item \textbf{Restoration.} A drop in certain key FoMs may signal that the most recent cull  degraded the model. 
This triggers restoration of the culled functional, as follows: 
 (i) restore the culled functional to the active library; (ii) add a flag to protect it from culling for the next few iterations. 
\end{enumerate}

\noindent \textbf{(iii) Cull a functional:} \\
``Culling a functional'' $f_i$ for $x_j$ means removing it from the active library of $x_j$, i.e. set $\xi_{ij} = 0$.
This stage operates on all active functionals combined, i.e. on the matrix $\Xi$. 
\begin{enumerate}
\setcounter{enumi}{8}
\item \textbf{Rescale the model coefficients} $\xi_{ij}$ according to the magnitude of their associated functionals $f_i$.
The rescaled coefficients are only used to decide which functional(s) to cull \ref{sectionWeightCoefficients}
\item For each $x_j$:
\textbf{To cull, first use linear dependence} among the active functionals (i.e. those $f_i$ with $\xi_{ij} \neq 0$). 
Run leave-one-out $L_2$ regression on the time-series of active $f_i$ to get $R^2$ values. 
If some $\{f_i\}$ are dependent (i.e. $R^2$ values above some threshold, e.g. 0.95), cull the one with the lowest rescaled $\xi_{ij}$, then skip the rest of culling for this iteration \ref{sectionCullViaLinearDependence}. 
If there are no linear dependence culls:
\item \textbf{Exclude some functionals  from culling}.
Active functionals (i.e. with non-zero $\xi_{ij}$) can be excluded from culling for two reasons:
(i) If they are protected due to recent restoration (\ref{sectionRestore}); 
(ii) We  optionally impose a \textbf{balance constraint} on the active functional counts of each variable $x_j$, so that functionals are not lopsidedly culled from only one variable. 
Exclude all $\xi_{*j}$ for any variable $x_j$ whose active functional count  is low enough to violate the balance constraint. \ref{sectionBalanceFunctionalCounts}
\item \textbf{Use rescaled $\xi_{ij}$ to cull functionals}, one per iteration \ref{sectionWeightCoefficients}, \ref{sectionOneCullPerIteration}
\item Save results of this iteration to file.
\end{enumerate}
\noindent Repeat iterations until all functionals are removed for some variable $x_j$, i.e. some row of the coefficient matrix $\Xi $ is zeroed out. 
Then:\\

\noindent \textbf{3. Final steps:}
\begin{enumerate}
\item (Optional) \textbf{Restart iterations} on the remaining variables, with new libraries, if there is noise variables are suspected \ref{sectionRestart}
\item \textbf{Choose the likely best model for each training trajectory} by consulting the FoM sequences (and possibly the text culling history) \ref{sectionChooseBestModels}
\item \textbf{Create a new library by taking the union} of the active $x_j$ libraries of each training trajectory's best model, i.e. $L = \{ f_i : \xi_{ij} \neq 0 \text{ for some } j \}$   \ref{sectionCombineBestModels}
\item \textbf{Repeat the full procedure} starting with this new library
\item Print new FoM mosaics, and choose an optimal model(s) \ref{sectionChooseBestModels}
\item \textbf{Assess non-uniqueness of solutions}: 
For each $x_j$, check (i) if certain functionals in the final sparse library are optional;  and (ii) if there are alternative candidate functionals in the span of the final sparse library. \ref{sectionFindInSpanFunctionals} 
\item Run the chosen optimal model(s) on test trajectories $(**)$
\item End of program.
\end{enumerate}

The next several subsections describe the various modules in more detail.
We suppose we have multiple trajectories of a dynamical system for training. 
If we have only one trajectory, the Figures of Merit based on validation trajectories cannot be used, but all other methods hold.





\subsection{Smoothing}
\label{sectionSmoothing}
\begin{figure}[t]
\centering
\centerline{
\fbox {
 \includegraphics  [width=0.6\linewidth]{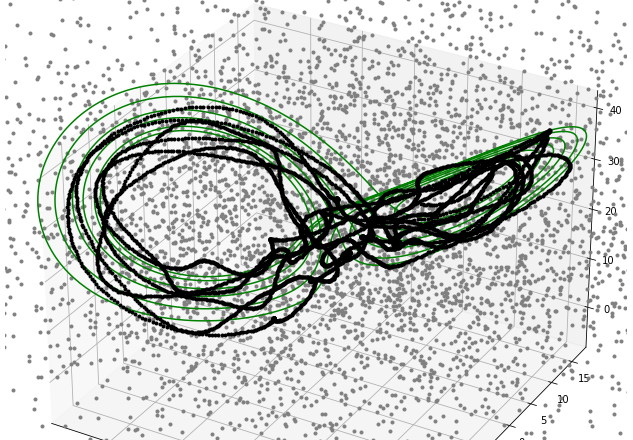} 
}
}
\caption{\small { \textbf{Squiggle artifacts introduced by low pass smoothing.} 
Lorenz trajectory with 150 - 200\% added noise. Green line= true clean trajectory, gray dots = noisy data, black line = smoothed trajectory.
Applying a low-pass filter (e.g. convolving with a Hamming window) creates squiggle artifacts in the smoothed trajectory.
These tend to derail standard SINDy methods but their effect is mitigated by other modules in the toolkit (using FoMs, weighting timepoints, and regressing on FFTs). 
}  }
\label{figNoisyAndSmoothedTraj}
\end{figure}

We smooth the noisy trajectory with a Low Pass filter (e.g. Hamming window). 
De-noising always requires parameter choices, and  introduces artifacts. 
In this case the parameter is Hamming window length, and the artifacts are squiggles due to noisy points within the window that distort the estimated trajectory (see Figure \ref{figNoisyAndSmoothedTraj}). 
These squiggles derail the fitting of standard SINDy models by distorting $\dot{x_j}$ and $f_i$ estimates.
With our toolkit, by contrast, the squiggles introduced by smoothing are rendered relatively harmless for three reasons:
(i) FoMs are relative to properties of the time-series $x_j$ (not the derivatives $\dot{x_j}$), and sparse models yield better FoMs due to their generalizing ability \ref{sectionFoms};
(ii) Weighting timepoints according to their estimated noise reduces the impact of large squiggles \ref{sectionWeightTimepoints};
(iii) Regressing on the FFT coefficients of the derivatives tends to neutralize the artifacts, since the high frequency  coefficients introduced by the squiggles have low magnitude and are thus sacrificed during $L_2$ optimization \ref{sectionFftRegression}.
The particular Hamming window length  is not critical as long as it is not much too big.

We note that other noise reduction methods 
could be used (in addition or instead) at this point, as long as they yielded continuous time-series $x_j(t)$.  


\subsection{Regression (estimating coefficients of active functionals)} 
Standard SINDy uses linear regression on the derivatives $\dot{x_j}$ of the system to estimate coefficients $\xi_{ij}$ of the functionals $f_i$.
Our toolkit offers several modifications to this basic program.


\subsubsection{Weighting timepoints for the target vector $y$}
\label{sectionWeightTimepoints}
All timepoints are weighted according to the \textit{z}-scores of their time-series values $x_j(t)$, as estimated from local noise envelopes, in order to downweight regression targets that are based on very noisy point estimates.
Each variable $x_j$ has a different vector of timepoint weights. 
Weights $w_{tj}$ are assigned to timepoint $t$ for $x_j$ as follows: 
\begin{enumerate}
\item Short windows of the time-series $x_j(t-n) ... x_j(t+n)$ are affine-transformed to have mean 0 (method: fit a line, then rotate and translate the points such that the line maps to the \textit{x}-axis 
\item The transformed values define a distribution
\item Interim weights $w_{interim}$ are defined as the inverses of the Mahalanobis distances (i.e. \textit{z}-scores) of the transformed  values, so that high \textit{z}-scores correspond to low weights
\item Log-scale the timepoint weights, $w$ =   ln$(w_{interim} + 1)$
\item The weight $w_t$ for timepoint $t$ is a combination of the weights $w_{sj}$ of the timepoints $s$ used to generate $\dot{x_j}(t)$ (e.g. the four timepoints $\{t-2, t-1, t+1, t+2\}$ in a Runge-Kutta approximation)
\end{enumerate}
This timepoint weighting has (we believe) the effect of favoring regression targets $\dot{x_j}(t)$ that depend on timepoints with low noise, which coincide with the sections of the smoothed trajectory with less squiggle.


\subsubsection{Ignore timepoints with extreme point values}
\label{sectionIgnoreExtremeFunctionalRatioPoints}
A second form of timepoint weighting (in fact exclusion), overlaid on the $\bm{w_j}$ above at each iteration, is based on the notion that sometimes the ratio between the values of library functionals  is so great that a regression will be unstable. 
For example, suppose one functional $f_1(t) \approx 10 \pm 2$, while a second functional $f_2(t) \approx 10sin(t)$. 
Most of the time the ratio $\frac{f_1}{f_2}$ is not extreme, but it blows up at $f_2$'s zero crossings. 
A regression using timepoints  near these zero crossings will likely result in much different, and more volatile, regression coefficients than a regression that uses only timepoints with  reasonably bounded $\frac{f_1}{f_2}$.
For this exclusion method, the ratios of an $x_j$'s active (i.e. not yet culled) functionals are calculated, and any timepoints with maximum ratios above some threshold (e.g. 30, exact value not relevant) are removed (weight = 0) from the regression target of that $x_j$.
This method serves to exclude the most pathological regions from the regression.


\subsubsection{Regress on the FFT  of $\dot{x}$ as well as on $\dot{x}$ }
\label{sectionFftRegression}
SINDy typically uses estimates of $\dot{x_j}$ as a regression target.
To these targets we add the coefficients associated with the derivatives' FFT.
If we regress on the complex FFT coefficients, this is in theory equivalent to regressing on $\dot{x_j}$ itself.
Alternatively, we can transgress mathematical correctness and use just the real part of the coefficients, the coefficient magnitudes, or the coefficients of the FFT power spectrum (this latter is in fact our default). 
Though none of these are strictly correct, they all serve to focus the regression on matching a characteristic of the derivatives distinct from the derivative values themselves.

In particular, the smoothed trajectory contains (i) lower frequencies associated with the trajectory's true behavior; and (ii) higher frequency artifacts (squiggles) created by the moving window filter.
We suspect that  regression on the FFT power spectrum coefficients knocks out the high frequency artifacts  because $L_2$ optimization minimizes its loss by preferentially failing on small-magnitude coefficients in order to match large-magnitude coefficients.
This pushes the regression to ignore the artifacts, reducing their harmful effect.


\subsubsection{Do multiple fits on subsets of timepoints (boosting)}
\label{sectionBoosting}
Especially in a noisy setting, the coefficient $\xi_{ij}$ for $f_i$ from regression on an $\dot{x_j}$ target vector is only a single draw from a distribution of possible coefficient values. 
In a greedy setting like STLSQ, an aberrant $\xi_{ij}$ can have catastrophic effects downstream if it leads to the cull of an important functional.
It also can cause inaccurate evolutions of train and val trajectories, and thus inaccurate FoMs.
Thus, at each regression we wish to estimate the distribution of  possible $\xi_{ij}$s  and choose the most representative option.
\cite{hirsh} developed a (computationally expensive) Bayesian method to estimate the distribution of each $\xi_{ij}$ . 
Here we apply the simpler and cheaper method of boosting.
We draw $n$ (e.g.15) random subsets $S$ (alternatively, sequential sections) of timepoints  and fit a model to each $S$. 
This yields a set of $n ~ \xi_{ij}$s for each functional $f_i$ (and $x_j$), from which to choose a suitable $\xi_{ij}$ (we take the median).
This method effectively controls the risk of aberrant $\xi$s at low cost.
This is ensembling at a local level (per iteration).
In concurrent, independent work, \cite{fasel} examines a variety of ensembling techniques for mitigating noise, applied to data and libraries at the model level.
 
\subsection{Figures-of-Merit based on model evolutions}

Standard SINDy judges a model based solely according to how well it fits $\bm{\dot{x}}$, i.e. each $\dot{x_j}$.
However, a good fit to $\dot{x_j}$ is usually a given and does not distinguish good models from bad.
Thus we wish to judge models according to whether their evolved time-series matches the given $x_j$s. 
To do this, we generate Figures of Merit (FoMs), which are various statistics that compare $\hat{x}_j(t)$ to the actual trajectory $x_j(t)$ (we have access to the noisy trajectory and its smoothed version).  
If we have multiple training trajectories $\bm{x}^k$, we create multiple models, where model $m^k$ is trained on $\bm{x}^k$ (its home trajectory) and the other trajectories $\bm{x}^{h \neq k}$ act as validation trajectories. 
To generate FoMs for  $m^k$, we evolve it on both its home and validation trajectories.

\subsubsection{Evolve the current model}
\label{sectionEvolve}

At each iteration, we   use an ODE solver that takes initial conditions (e.g. Python Scipy's solve\_ivp or odeint methods) to generate  a prediction $\bm{\hat{x}}(t)$ for the home and for each validation trajectory. 
Initial conditions for the evolution are found via weighted averaging of a small neighborhood of points around the nominal starting point, with more trustworthy points weighted heavier (cf section \ref{sectionWeightTimepoints}).
This matters in chaotic systems where small changes in initial condition can strongly affect the evolution.

We note that  these evolutions are the most computationally costly step of the toolkit, especially if a particular model has points of stiffness, or if we are tracking model stability (since we need to do multiple evolutions to see if these diverge). 
This cost can be mitigated in three ways: 
(i) By parallelization, since the different train-val splits can run in parallel, and evolutions of a particular model at each iteration can also run in parallel;
(ii) By doing only one evolution, if stability of a model is not in doubt;
(iii) By skipping evolutions in early iterations when active libraries are still large and FoMs tend to be uninformative.
 
\subsubsection{Figure-of-Merit histories}
\label{sectionFoms}
  
\begin{figure}[t]
\centering
\centerline{
\fbox {
\includegraphics  [width=1 \linewidth]{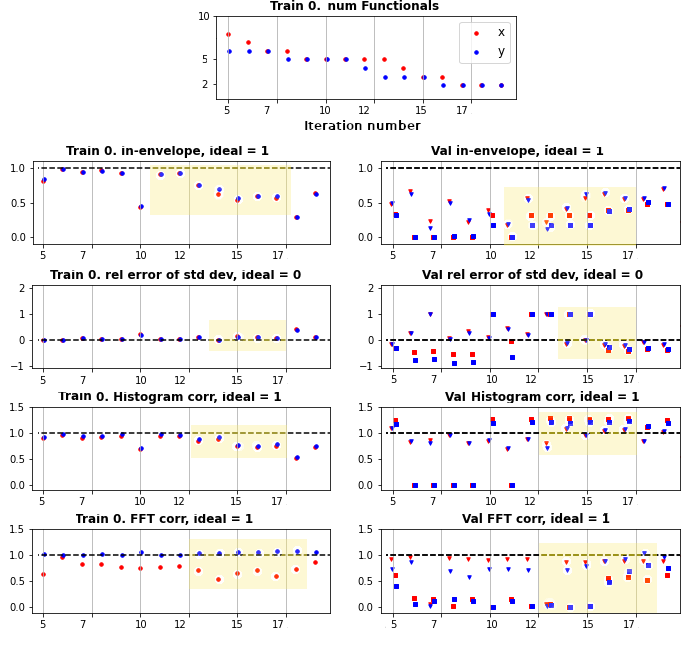} 
}
}
\caption{\small { \textbf{Figures of Merit for train and val trajectories.} 
Harmonic cubic oscillator; true model has 2 functionals per variable.
Plots show useful FoMs, and how validation trajectory FoMs add information. 
Relevant sections noted below are highlighted in yellow.
Dotted lines show ideal values.
{\bf{Row 1:}} The number of functionals for  $x$ and $y$ in the model, as sparsity increases through iterations. 
{\bf{Left Column, Rows 2--4:}} FoMs for the Home (training) trajectory. 
{\bf{Right Column, Rows 2--4:}} FoMs for the two Validation trajectories (one trajectory as squares, one as triangles). 
{\bf{Row 2:}} Fraction of the evolved trajectories ``in envelope''.
Note the trade-off in ''in envelope'' FoM between Home trajectory and Validation trajectories after iteration 12: training accuracy decreases, but validation accuracy increases.
{\bf{Row 3:}} Relative error of std dev of evolved trajectory (0 is ideal). 
Note that the Home accuracies are consistently good while Validation accuracies vary substantially by iteration.
{\bf{Rows 4 and 5:}} Correlation to histograms of evolution values (row 4) and FFT power (row 5). 
In both cases, Home accuracies are generally high, while Validation accuracies vary by iteration.
In row 4,  Home accuracy drops after iteration 14 while Validation accuracy remains high.
In rows 4 and 5 of the Validation (right-hand) column, some models show good fit to one trajectory (triangles) but not the other (squares).
}  }
\label{figFomMosaicFocused}
\end{figure}

For each training trajectory $\bm{x}^k$, iterative culling of functionals gives a sequence of progressively sparser models $m^k$.
FoMs are recorded for each model in the sequence.
These FoMs are then plotted vs iteration number as shown in Fig \ref{figFomMosaicFocused}, allowing the user to select which among the models best matches the desired time-series behavior.
This method (do one complete STLSQ run and record the models at each step) is similar to how Lasso is done, and is an alternative to sweeping the sparsity parameter $\lambda$ with multiple complete SINDy runs.
The model at each iteration is also printed to text file for later inspection.
 
Examples of FoMs include: 
\begin{enumerate}
\item Whether multiple evolutions match each other. 
This assesses stability of the model.
\item Fraction of evolved trajectory within upper and lower bounds (of the true trajectory). 
This detects egregious blow-ups.
\item Fraction of evolved trajectory within an envelope about the true trajectory. This is a finer measure of whether the evolved trajectory tracks the true, and is very valuable.
\item Relative error of the std dev of evolved trajectory $v(t)$ vs. true trajectory $x(t)$, $\frac{\sigma(v) - \sigma(x)}{\sigma(x)}$. 
This comparison of std devs is a very valuable measure. 
\item FFT power spectrum correlation between evolved and true trajectories.
\item Histogram correlation between evolved and true trajectories.
\end{enumerate} 
 

\subsection{Culling functionals}
STLSQ involves culling functionals with the lowest  non-zero coefficients in the estimate $\hat{\dot{x}}_j = \Sigma \xi_{ij} f_i$ where the active (i.e. not yet culled) functionals for $x_j$ are those with non-zero $\xi_{ij}$.
The toolkit offers several modifications to this sequential culling:


\subsubsection{Cull one functional per iteration}
\label{sectionOneCullPerIteration}
We wish to capture the importance of a given functional's loss in the FoM sequences.
To tie individual functionals to the FoMs, we cull just one functional (from all variables combined) per iteration.
We continue culling until one variable loses all its functionals (with an optional restart,  see \ref{sectionRestart}).
A faster alternative, but  with less granularity in terms of FoMs, would be to cull multiple functionals per iteration. 
A workable compromise is to cull multiple functionals during early iterations, then to reduce to one functional per iteration later as the active libraries get sparser.


\subsubsection{Cull via linear dependence of functionals}
\label{sectionCullViaLinearDependence}
\begin{figure}[t]
\centering
\centerline{
\fbox {
\includegraphics  [width=1\linewidth]{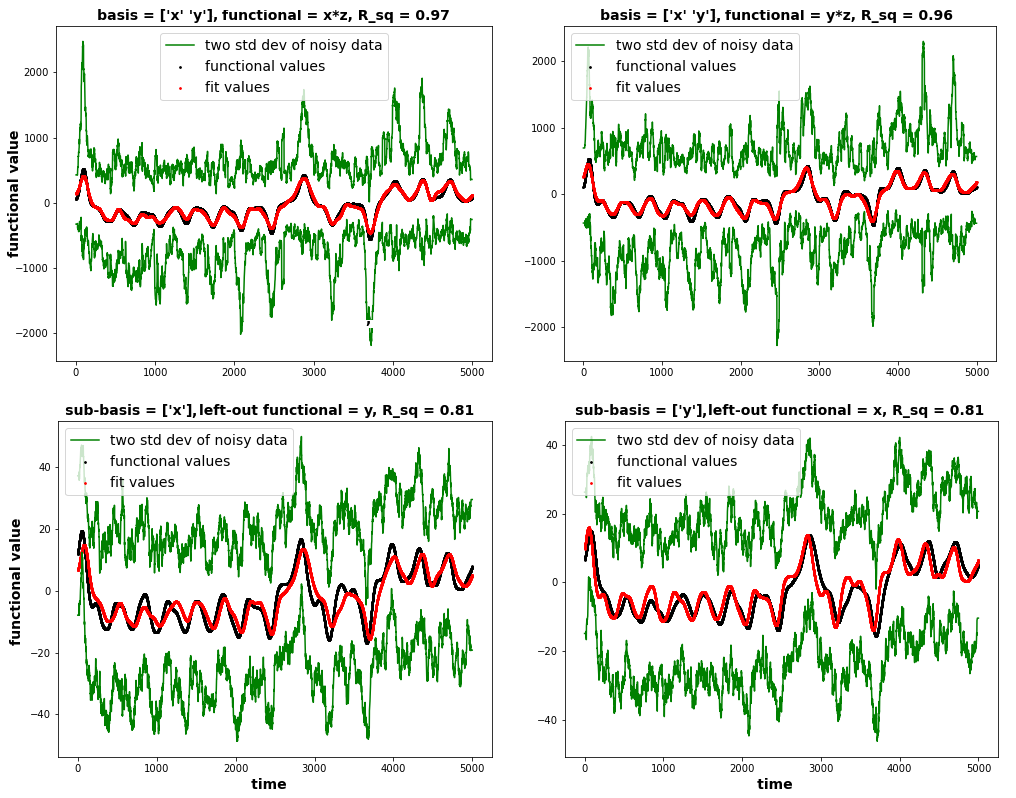} 
}
}
\caption{\small { \textbf{Linear dependence of functionals.}  
Lorenz system.
For each subplot: \textit{x}-axis = time, \textit{y}-axis = functional value.
Black = functional to be fitted (values using smoothed data), red = linear fit by basis functionals, green = noise envelope (2 std dev). 
{\bf{Top Row: }}Given a discovered sparse set of governing equation functionals, we can test whether other (culled) functionals are in their linear span, relative to the noise envelopes.  
If yes, they are potential candidates for the ``true'' governing equations even though they were culled.
In Lorenz's $\dot{x}$, $xz$ and $yz$ appear to be within the span of the discovered set \{$x$ and $y$\} ($R^2  \approx 0.97$
), so they are plausible``true'' functionals, perhaps instead of $x$ or $y$.
{\bf{Bottom Row:}} We can also test via leave-one-out whether any of the discovered active functionals are redundant.
In Lorenz's $\dot{x}$, the selected functionals $x$ and $y$ have relatively poor leave-one-out fits ($R^2$ = 0.81), indicating they are both essential (linearly independent).
}  }
\label{figLinDepLorenzX}
\end{figure}

Various functionals in a library may be de facto (i.e. given the noise level) linearly dependent, with an ``incorrect'' functional  in the span of a ``correct'' functional.
 For example,  in Lorenz, $\dot{x} = \xi_1 x + \xi_2 y$, but $x z$ and $yz$ are in the span of $x$ and $y$ (see Figure \ref{figLinDepLorenzX}).
In this case, the coefficients chosen by $L_2$ regression are but one choice among many, and other linear combinations of functionals are effectively equivalent.

We address this as follows:
Each iteration, we check the linear dependence of the active functionals (per each variable separately), calculating $R^2$ values for leave-one-out $L_2$ fits.
If some functionals show linear dependence via high $R^2$ values (above some threshold e.g.0.95), we cull the one with the lowest rescaled (cf \ref{sectionWeightCoefficients}) coefficient.
That is,  we cull based on coefficient magnitude (as usual), but we restrict that iteration's culling candidates to just those functionals that are linearly dependent. 
This method handles large initial libraries well, reliably culling excess functionals.

\subsubsection{Rescale the functional coefficients according to magnitude of functional values}
\label{sectionWeightCoefficients}
Library functionals often have radically different value ranges, which can strongly bias culling based on coefficient magnitude.
For example, suppose $x(t)$ is generally around 10. 
Then $x \approx 10, x^2 \approx 100$, etc.
This translates directly into much smaller coefficients $\xi$ for some functionals, which increases the likelihood they will be culled by thresholding, regardless of whether they are part of the ``true'' governing equation.
To mitigate this bias, we cull coefficients not based on their regression coefficients $\xi$, but on rescaled versions $v_i\xi $ such that the  contribution of each $v_i \xi_i  f_i$ term is (very roughly) equal.
We approximate this ideal for a particular $x_j$ as follows: 
For each active functional we calculate a median value over the fitted timepoints (=$med(f_i)$), then normalize by some percentile of all active functionals' medians to get rescaling factors $v$:

$v_i = \frac{med(f_i)}{M}$  where  $M = m^{th}$ percentile of \{$med(f_i)\}$ over all active $f_i$.

The parameter $m$ determines which active functional is considered ``standard'', i.e. has $v = 1$.
Then $v_i \xi_i > \xi_i$ if $med(f_i) > M$ (relatively high values of $f_i$ lead to relatively low values of $\xi_i$, which $v_i$ offsets).
Similarly, $v_i \xi_i < \xi_i$ if $med(f_i) < M$ (relatively low values of $f_i$ lead to relatively high values of $\xi_i$, which $v_i$ offsets).

The $v_i$ depend on which $f_i$ are active ($\xi_{ij} \neq 0$), so they differ for each $x_j$ and also change as functionals are culled.
This rescaling method replaces the ``do-nothing'' default  that benefits small-valued functionals with a deliberate choice that balances large- and small-valued functionals by increasing (for culling purposes only) the coefficients of large-valued functionals. 


\subsubsection{Constrain the imbalance in coefficients per variable}
\label{sectionBalanceFunctionalCounts}
STLSQ, if applied to the full system (ie all variables $\dot{x_j}$) removes the functional with smallest coefficient, regardless of which variable that functional acts on.
This can lead to imbalances, where one variable retains a dense library of active functionals while another variable becomes overly sparse. 
Essentially, the complexity in the overall system is incorrectly captured by one variable's library.
A domain expert might have educated guesses about the relative symmetry of a system.
We encode this as a constraint on the maximum allowed difference in library sizes between variables (e.g. 3). 
If in a particular iteration one variable has many fewer active functionals, those functionals are ignored by the culling.
In this case some other variable losing a functional, bringing the total counts into closer balance. 


\subsubsection{Restore and protect culled functionals if FoMs drop}
\label{sectionRestore}
A basic problem with greedy algorithms such as STLSQ  is that a ``true'' functional may get culled early and is then permanently lost, which hurts the performance of downstream models.

\begin{figure}[t]
\centering
\centerline{
\fbox {
\includegraphics  [width=1\linewidth]{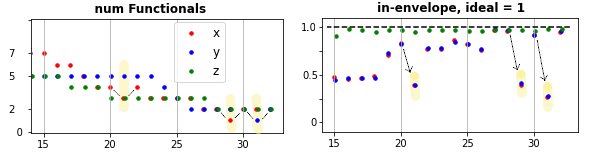} 
}
}
\caption{\small { \textbf{Restoring culled functionals.}
Large degradation of the ``in-envelope'' FoM due to culling a functional triggers restoration of the culled functional.
In the 3-D linear harmonic oscillator, the ``true'' functionals for $\dot{x}$ and $\dot{y}$ are $x$ and $y$.
At iteration 21 $x$ was culled from the active functionals for $\dot{x}$ (Left), causing degradation of the ``in-envelope'' FoM (Right). 
This triggered restoration of $x$ at iteration 22 and temporarily placed it under protection from culling.
The two incorrect functionals were subsequently culled from $\dot{x}$ leaving only $x$ and $y$. 
At iteration 29, $x$ was again culled from $\dot{x}$, degrading the ``in-envelope'' FoM and triggering restoration.
Similarly, at iteration 31 $y$ was culled from $\dot{y}$ causing degradation, triggering its restoration.  
}  }
\label{figCullDegradesFomThenRestore}
\end{figure}

Because we collect FoMs at each iteration, we have immediate notice if culling a particular functional degrades model performance. 
Sufficient degradation (e.g. 50\% reduction in some FoM for some variable) triggers (i) restoration of the culled functional, and (ii) protection of that functional from culling for the next few iterations.
This allows time for other functionals to be culled instead, which changes the landscape of coefficient values.
If the restored functional's coefficient increases, it gets preserved going forward; while if its coefficient remains low, it gets culled later.
See Figure \ref{figCullDegradesFomThenRestore} for an example of restoration.
 
 
\subsection{Final steps}

\subsubsection{Restart process on remaining variables}
\label{sectionRestart}
The culling iterations continue until all functionals have been removed for one variable.
The procedure can optionally restart the STLSQ iterations on the remaining variables.
The new iterations use the original full library of functionals, minus any functionals containing the removed variable.
This is an effective way to detect and remove pure noise variables whose $\dot{x}$ = constant.
It is an irrelevant step if all the original variables were salient.

\subsubsection{Choose best models}
\label{sectionChooseBestModels} 
 By consulting an FoM mosaic, one can select an optimal model that performs well on both Train and Validation trajectories, and is sufficiently sparse. 
In addition, the FoM sequences give clues as to whether a potentially relevant functional was incorrectly dropped early, since this often causes a  drop in  FoMs in the iteration when it was culled. 
Validation FoMs give  insight into generalizability.
The text print-out of model coefficients at each iteration can be matched with the FoM mosaics to glean insights into which functionals may be most important. 
Also,  select models can be evolved over train and validation trajectories, to allow visual inspection of their predictions.
 
\subsubsection{Combine best models}
\label{sectionCombineBestModels} 
If there are multiple training trajectories, each trajectory produces a sequence of models and an FoM mosaic, and thus a different optimal model.
These models usually  have different sparse libraries due to differences in training trajectories and vicissitudes of execution.
In this case, the active functionals of the various models can be combined (a union of sparse libraries), and the full procedure restarted on all trajectories with this new initial library (which is much smaller).
Note that the union of libraries respects the differences between variables: For example, if  some model has $\xi_{i1}\neq 0$, but all models have $\xi_{i2} = 0$, then in the union library $f_i$ is included for $x_1$ but not for $x_2$.
Functionals suspected of having been incorrectly dropped (based on a drop in FoMs at some iteration) can also be reinstated at this point.

This method usually  improves the discovered models substantially.
It  combines the positive findings of each training trajectory to create a concentrated starting library of highly-likely candidate functionals.
It also  mitigates greedy culls of ``true'' functionals, since the loss is reversed if the functional was preserved by another trajectory.


\subsubsection{Find alternate acceptable library functionals}
\label{sectionFindInSpanFunctionals}
As mentioned, a set of functionals may be effectively linearly dependent, $f_k(t) \approx \Sigma_{i \neq k} \beta_i f_i(t)$, if the approximation is well within the noise envelope of the data.
Thus there may be multiple plausible sparse models for a system. 
Given a final model, we do two types of checks for linear dependence (examples are from the Lorenz system with 150 - 200\% added noise):  

\begin{enumerate}
\item We apply linear regression to each culled functional's time-series, using the retained functionals' time-series as features. 
The goodness of fit, e.g. shown by $R^2$ values,  indicates whether the culled functional might be a viable alternative to the chosen functionals for defining governing equations. Examples:\\ 
(i) $\dot{x} = \beta_x x + \beta_y y$; but $xz \approx \beta_1 x + \beta_2 y$ ($R^2$ = 0.97) and $yz \approx \beta_3 x + \beta_4 y$ ($R^2$ = 0.96) (see Figure \ref{figLinDepLorenzX} row 1).
So $xz$ and $yz$ are otentially  viable substitutes for $x$. \\
(ii) $\dot{y} = \beta_x x + \beta_y y + \beta_{xz} xz$.
But  $yz \approx \beta_1 x + \beta_2 y + \beta_3 xz ~(R^2 = 0.99)$, so $yz$ is a viable substitute for one of the ``true'' functionals.
\item We apply ``leave-one-out'' linear regression to the set of retained functionals, fitting each functional's time-series using the time-series of the other functionals. 
This gives clues as to whether the retained functionals are all necessary. Examples: \\
(i) $\dot{x} = \xi_x x + \xi_y y$, and neither can be well approximated by the other ($R^2 \approx 0.81$, see Figure  \ref{figLinDepLorenzX} row 2). 
So they are both essential. \\
(ii) $\dot{y} = \xi_x x + \xi_y y + \xi_{xz} xz$. 
However, $x$ or $xz$ might be redundant:
$x \approx \beta_1 y + \beta_2 xz$ ($R^2$ = 0.98) and $xz \approx \beta_3 y + \beta_4 x$ ($R^2$ = 0.97).
But $y$ is not in the span of $x$ and $xz$ ($R^2 \approx 0.89$), so it is not redundant.
 ``Leave-one-out'' in-span behavior is not necessarily symmetric.
\end{enumerate}

This method acknowledges that solutions may not be unique given high-noise data. 
The output is a list of possible alternative functionals that are in the linear span (in the sense described above) of the discovered functionals.
This list of alternatives enables domain experts to identify other functionals potentially viable for use in governing equations, rather than being constrained to just a single discovered model.
 

\subsection{Assessment of discovered models using linear dependencies}
\label{sectionLinearSpanAssessment}

In the previous section,  linear dependence was used to help domain experts identify alternate candidate functionals.
In this section, the goal is oracle assessment of whether a discovered model has an equivalent form that is close to the given ground truth.
This  problem and technique are not specific to SINDy, but apply to data-driven equation discovery in general given noisy data.

Our ability to assess whether a discovered model matches the original ``true'' model is complicated by the non-uniqueness of solutions  in noisy regimes:
There may be multiple sparse coefficient matrices $\Xi$ such that $\bm{\dot{x}} = \Xi F$ accurately reproduces dynamics of the system. 
We describe construction of a non-degenerate linear transform  that converts the discovered model coefficients $\hat\Xi$ to a new, equivalent form $\hat\Xi '$ which as closely as possible matches the ``true'' $\Xi$, while maintaining the same dynamical behavior.
Our goal is to accurately assess errors in discovered sparse functional libraries and in estimated coefficients.
For this purpose we assume oracle knowledge of the ``true'' model.
The method is as follows:

\begin{enumerate}
\item Take as features the time-series of the ``true'' model's  library $L$ over a set of timepoints $T$.
For each $x_j$, do linear fits of each  functional $g(t)$ in the discovered model, $g(t) = \Sigma_{i \in L} \beta_i f_i(t) + \epsilon$, and record the $R^2$ value of the fit (cf section  \ref{sectionFindInSpanFunctionals}).
\item Based on a pre-set $R^2$ threshold (e.g. 0.95) that indicates a sufficiently close fit, see if any $g$ are in the span of the true functionals. 
This threshold partly depends on the size of the noise envelopes. 
\item If a discovered $g$ is in the span of $L$, use the linear relationship to substitute $g$ out of  the discovered model, replacing it with true functionals.
After this step, the transformed sparse library overlaps the ``true'' sparse library as closely as the $R^2$ threshold allows.
\item Using leave-one-out linear fits, calculate the linear dependencies within the transformed library. 
\item If there are linear dependencies with sufficiently high $R^2$ values, apply iterative substitutions to shrink the largest  error in the modified vs ``true'' coefficients $|\frac{\hat{\xi}_{ij} - \xi_{ij}}{\xi_{ij}} |$.
After these iterations, the discovered model has a form whose coefficients match the ``true'' model as closely as the $R^2$ threshold allows, in the sense of having smallest maximum error.
A side effect of minimizing errors of true functionals is to minimize the coefficients of surplus functionals.
\item Evolve the transformed model, to confirm that trajectories are  unchanged (or improved).
\end{enumerate}

\begin{figure}[t]
\centering
\centerline{
\fbox {
 \includegraphics  [width=1\linewidth ]{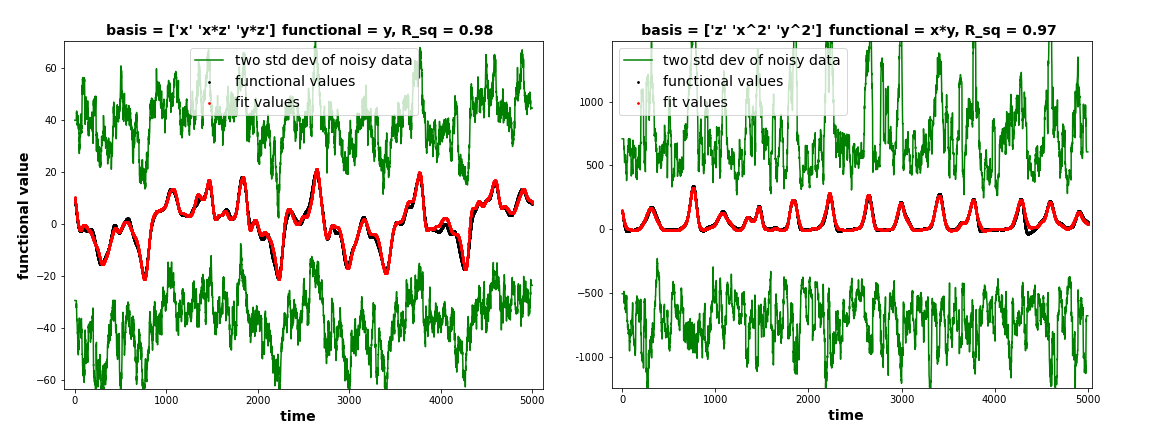} 
}
}
\caption{\small { \textbf{Culled ``true'' functionals are often in the linear span of discovered models.}  
From the discovered model for one of the training trajectories, Lorenz with 220 - 300\% added noise.
{\bf{Left}}: The discovered sparse library for $\dot{x}$ was $\{x, xz, yz\}$. 
The ``true'' functional $y$ was culled, but  was in the span of the discovered functionals ($R^2 \approx 0.98$).
{\bf{Right}}: Similarly, the discovered model for $\dot{z}$ was $\{x$, $x^2, y^2\}$.
The true functional $xy$ but was in the span of the discovered functionals ($R^2 \approx 0.97$).
}  }
\label{figLorenzHighNoiseLinDepOfTrue}
\end{figure}

\begin{figure}[t]
\centering
\centerline{
\fbox {
\includegraphics  [width=1\linewidth]{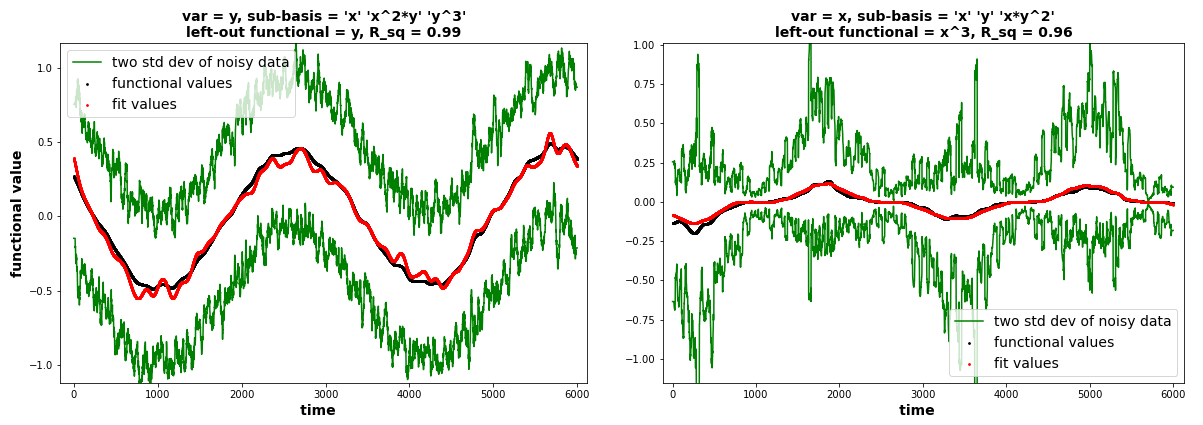} 
}
}
\caption{\small { \textbf{Linear dependence of functionals, Hopf Normal  2-D.} 
For each subplot \textit{x}-axis = time, \textit{y}-axis = functional value.
Black = functional to be fitted (values using smoothed data), red = linear fit by basis functionals, green = noise envelope (2 std dev).  
Each plot shows the linear dependence of a left-out functional (left: $y$, right: $y^3$) on the other ``true'' functionals of $\dot{y}$, in the form of a linear regression with $R^2$ > 0.95.
An extreme example is shown on the left: $y = 0.02x + 4.34y^3 + 4.54x^2y \implies 0 = -y + 0.02x + 4.34y^3 + 4.54x^2y$ with $R^2$ = 0.99. 
These equalities can be substituted in to find a version of the discovered model that is closest in form to the ``true'' model. 
}  }
\label{figHopf2DLinearDeps} 
\end{figure}

Examples: Fig \ref{figLorenzHighNoiseLinDepOfTrue} shows missing ``true'' functionals that are linear combinations of discovered functionals.
Fig \ref{figLinDepLorenzX} shows linear dependencies as used in step 3 above.
Fig \ref{figHopf2DLinearDeps} shows linear dependencies as used in step 5 above.  
Results of oracle transformation are evident in Tables \ref{tableResults50} to \ref{tableResults300}, and Tables \ref{tableHarmCubOsc70} and \ref{tableHopf2DModelsErrors70}.

%





\section{Results}

We give results for several dynamical systems with added noise: Lorenz, linear and cubic harmonic oscillators, linear 3-D, and the Hopf Normal form (2-D). 
Background on these systems can be found in \cite{Brunton2016pnas}.
The toolkit performs well on all these systems at varying levels of noise.
Results for the Lorenz system are given in this section.
Results for the other systems are in the Appendix.
The Lorenz system (here) and Hopf system (Appendix) clearly show the importance of the linear dependence method (cf \ref{sectionLinearSpanAssessment}) when assessing discovered models. 

While the toolkit requires various hyperparameters, none of them are brittle, and a wide range of values work well (perhaps because there are so many relatively ``easy'' gains to be had). 
The Lorenz system was our test-bed, which perhaps partially explains the higher levels of noise handled for Lorenz. 
The other systems were handled using the same hyperparameters used for Lorenz, with no additional tuning.

\subsection{Added noise}
In these experiments, white noise was added as follows:
Each variable $x_j(t)$ of a trajectory was transformed by FFT to give complex frequency coefficients. 
Noise drawn from a complex Gaussian distribution $\mathcal{N}(0, \sigma)$ was added to each coefficient. 
An inverse Fourier transform gave a complex-valued trajectory, of which the real part was retained. 
Noise level (\%) was defined as $100\times~\sigma_{noise} / \sigma_{x}$ where $\sigma_{noise}$ was standard deviation of the noise-added trajectory minus the clean trajectory, and $\sigma_x$ was standard deviation of the original clean trajectory. 


\subsection{The Lorenz system}
\label{sectionResultsLorenz}

We consider the canonical Lorenz system, a 3-dimensional chaotic ``butterfly''-shaped attractor with two lobes, shown previously in Fig \ref{figIncreasingNoiseLorenz} (time-series view) and also below in Fig \ref{figLorenzEvolvedTestIncreasingNoise}  (3-D view).  
White noise was added at levels equivalent to 50\%, 100\%, 200\%, and 300\%, and results for typical runs are reported.

The true system has ODEs:
\begin{flalign}
&\dot{x} = -10x + 10y   \\
&\dot{y} = ~~~28x - y  - xz   \\
&\dot{z} = -2.67z + xy   
\end{flalign}

We used three training trajectories with initial conditions [-8, 8, 27], [5, -7, 29], and [-2, 7, 21]; and two holdout trajectories with initial conditions [8, 7, 15] and [-6, 12, 25].
The initial conditions of the first training trajectory are from \cite{Brunton2016pnas}; all others were selected at random.
The training trajectories were 10 seconds long with 0.002 second timestep.
The initial functional libraries consisted of all polynomials up to degree 4 (almost all the degree 4 polynomials were rapidly eliminated by culling based on linear dependence, cf section \ref{sectionCullViaLinearDependence}).

The method typically recovers the correct sparse functional libraries (or, in some cases, linearly dependent equivalent libraries).
However, as noise increases the coefficient estimates become less accurate, which leads to worse prediction on holdout trajectories. 
If the coefficient error becomes large enough, the qualitative behavior of predicted trajectories changes.

For results at all noise levels, the discovered and ``closest to true'' models (three models, one per training trajectory) are listed along with coefficient errors. 
The key is as follows:
Column 2 gives the raw discovered equations.
Column 3 gives the ``closest to true'' equations, after transformation using linear dependencies with $R^2 \geq 0.95$.
Columns 4 and 5 give the absolute coefficient errors for each true functional, $ |  ( \hat{\xi} - \xi )  /  \xi   |$ as percentage (or ``\textit{inf}'' if the functional is missing).

In the true library of $\dot{y}$, the  $y$ functional is linearly dependent on the $x$ and $xz$ functionals, so it could be substituted in, as seen in the transformed (``closest to true'') versions of  $\dot{y}$.
In $\dot{x}$, some incorrect functionals were in the span of the true library (cf Fig \ref{figLorenzHighNoiseLinDepOfTrue}), allowing  substitution in some cases.
The true library of $\dot{z}$ had no relevant linearly dependencies, so no coefficient transforms were possible. 

Which discovered model would give the best test set evolutions was in all cases fully predictable based on quality of training and especially validation set predictions.


\subsubsection{50\% added noise}
\label{sectionResultsLorenz50Noise} 
 
\begin{figure}[t]
\centering
\centerline{
\fbox {
 \includegraphics  [width=1\linewidth ]{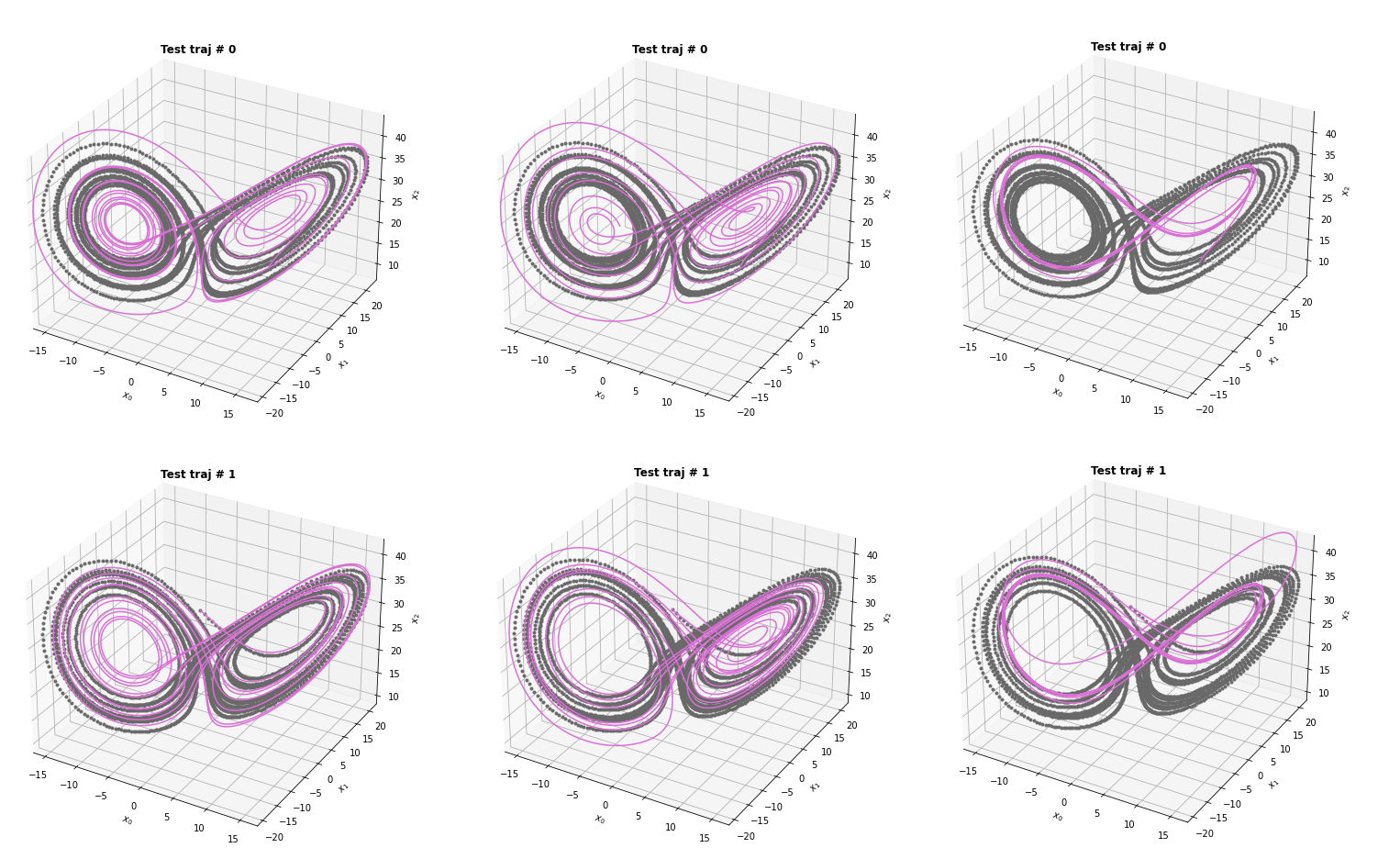} 
}
}
\caption{\small { \textbf{Test set evolutions for 50\%, 100\%, and 200\% training data noise.} 
Evolutions on two test trajectories of models trained on data with 50\% to 200\% added noise (cf Figure \ref{figIncreasingNoiseLorenz}).  
Evolutions are from the model with the best train and validation trajectory predictions.
Grey dots: original test data. Purple lines: Evolved trajectory.
Evolved trajectories are good at lower noise, and deteriorate somewhat as noise increases, though overall trend  and qualitatively correct behavior is maintained.
At 300\% noise (not shown), the predicted test trajectories degenerate into simple figure-8s (i.e. one loop per lobe, rather than multiple loops in a lobe before switching).
{\bf{Col 1:}} Noise level 50\%. 
{\bf{Col 2:}} Noise level 100\%. 
{\bf{Col 3:}} Noise level 200\%. 
{\bf{Top row:}} Test trajectory 0. 
{\bf{Bottom row:}} Test trajectory 1.
}  }
\label{figLorenzEvolvedTestIncreasingNoise}
\end{figure}

Typical discovered models and coefficient errors for 50\% added noise  are given in Table \ref{tableResults50}.
In general, correct functionals (an exception is given below) were selected, with very low coefficient error.
In all models, the raw $\dot{x}$ and $\dot{z}$ equations had ``correct'' functionals. 
All models missed the $y$ functional in the $\dot{y}$ estimate. 
However, evolutions of validation and holdout trajectories were still very accurate, which highlights the subjective nature of assessing ``correctness'' in the model discovery context.
Test set predicted evolutions of the best model (determined on train-val results) are shown in Fig \ref{figLorenzEvolvedTestIncreasingNoise} (left column).

\begin{table}[t]
  \begin{center}
    \caption{Lorenz \textbf{50}\% noise: 
Discovered  (raw) and equivalent models, and absolute coefficient errors for each true functional, $|( \hat{\xi} - \xi )  /  \xi |$ as percentage (``\textit{inf}'' indicates a missed true functional). 
``Raw'' errors are for the discovered equation, ``closest'' errors are for the transformed equation (cf section \ref{sectionLinearSpanAssessment}).
}
\label{tableResults50}
    \begin{tabular}{c|c|c|l|l} 
      \hline
   \rule{0pt}{2.0ex}     ODE   & Raw Eqn   &  Closest Eqn &  Raw Err \%  & Closest Err \%  \\  
 \hline
 Model 0& & & & \\
      $\dot{x}$  & $-10.13 x  + 10.16 y$       &    same    &  (1, 2) &   (1, 2)  
  \\
      $\dot{y}$ & $~24.39 x   -0.93 xz$ & $~27.73 x   -1.01 y  -1.01 xz$ & (13, \textit{inf}, 7)   &  (1, 1, 1)   \\
      $\dot{z}$ &  $-2.62 z  + 1.02 xy $  &   same  & (2, 2)    &  (2, 2)     \\ 
\hline
 Model 1& & & & \\
      $\dot{x}$  & $-9.8 x  + 9.91 y$            &  same   &  (2, 1) &   (2, 1)  
  \\
      $\dot{y}$ & $~0.8 + 24.04 x   -0.92 xz$ & $~0.8  + 27.29 x  -1.01 y   -1.0 xz$ & (14, \textit{inf}, 8)   &  (3, 1, 0)  \\
      $\dot{z}$ & $ -2.65 z  + 1.04 xy$   &  same   & (1, 4)   & (1, 4)   \\
 \hline
 Model 2& & & & \\
      $\dot{x}$  & $-9.72 x  + 9.7 y$            &   same  &  (3, 3) &   (3, 3)  
  \\
      $\dot{y}$ & $~25.48 x  -0.97 xz$  & $~27.66 x -0.65 y  -1.02 xz$ & (9, \textit{inf}, 3)  & (1, 35, 2)   \\
      $\dot{z}$ & $-2.58 z  + 1.02xy  $ &   same  & (3, 2)    &  (3, 2)     \\
\hline
    \end{tabular}
  \end{center}
\end{table}


\subsubsection{100\% added noise}
\label{sectionResultsLorenz100Noise}
 
Typical discovered models and coefficient errors for 100\% added noise  are given in Table \ref{tableResults100}.
Raw models 0, 1 gave simple figure-8 trajectories, while raw model 2 gave more qualitatitively accurate trajectories.
In all cases, the transformed models gave improved, qualititatively realistic trajectories and discovered models contained the correct functionals: 
The raw $\dot{x}$ equations used $xz$ (``wrong'') instead of $x$, but linear dependencies gave equivalent forms with the ``true'' functionals $x$ and $y$.
The raw $\dot{y}, \dot{z}$ equations contained the true functionals. 
Test set predicted evolutions of the best model (based on train-val results) are qualitatively correct, as shown in \ref{figLorenzEvolvedTestIncreasingNoise} (center column).

Transforming using $R^2$ threshold = 0.85 (vs 0.95) gave models with (a) lower errors in coefficient estimates (median errors = 9, 6, and 4\%); (b) trajectories that were improved vs raw but were not quite as good as when using a 0.95 threshold.
These results highlight that care is required choosing an $R^2$ threshold.
We propose two  criteria: Closely matching regressions sitting well within the noise envelope; and similar evolutions on train-val trajectories. 

\begin{table}[t]
  \begin{center}
    \caption{Lorenz \textbf{100}\% noise: 
Discovered  (raw) and equivalent models, and absolute coefficient errors for each true functional, $|( \hat{\xi} - \xi )  /  \xi |$ as percentage (``\textit{inf}'' indicates a missed true functional). 
``Raw'' errors are for the discovered equation, ``closest'' errors are for the transformed equation (cf section \ref{sectionLinearSpanAssessment}) using $R^2$ threshold 0.95.
}
\label{tableResults100}
    \begin{tabular}{c|c|c|l|l} 
      \hline
    \rule{0pt}{2.0ex}    ODE   & Raw Eqn   &  Closest Eqn &  Raw Err \%  & Closest Err \%  \\  
 \hline
 Model 0& & & & \\
      $\dot{x}$  & $0 x + 5.88 y  - 0.18 xz$       &   $-6.9 x  + 7.55 y $  & (\textit{inf},  41)  &   (31, 24)   \\
      $\dot{y}$ & $21.37 x  + 1.93 y   -0.96 xz$ & $26.43 x  + 0.51 y   -1.08 xz$ & (24, 293,   4)   &  (6,  151 ,   8)   \\
      $\dot{z}$ &  $-2.52 z  + 1.01 xy $  &   same  & (6, 1)    &  (6, 1)     \\ 
\hline
 Model 1& & & & \\
      $\dot{x}$  & $0 x + 5.29 y   -0.14 xz$            &  $5.32 x  + 6.53 y $ & (\textit{inf},  47)&  (47,  35)   \\
      $\dot{y}$ & $20.19 x  + 2.25 y   - 0.89 xz $ & $26.96 x  + 0.42 y   -1.06xz$ & (28,  325,   11)    &  (4, 142,  6) \\
      $\dot{z}$ & $ -2.62 z  + 1.06 xy$   &  same   & (2,   6)   & (2,   6)  \\
 \hline
 Model 2& & & & \\
      $\dot{x}$  & $0 x + 6.48 y  + -0.22 xz$            & $-8.37 x  + 8.67 y $&  (\textit{inf},  35)&   (16,  13)   \\
      $\dot{y}$ & $~20.12 x  + 1.83 y   -0.83 xz$  & $~27.81 x  - 0.45 y   - 1.01xz$ &(28,  283,   17)  & (1,  55,  1)   \\
      $\dot{z}$ & $-2.49 z  + 0.99xy  $ &   same  & (7,  1)    &  (7,  1)     \\
\hline
    \end{tabular}
  \end{center}
\end{table}


\subsubsection{200\% added noise}
\label{sectionResultsLorenz200Noise} 

Added noise was 205\% to 220\%.
Typical discovered models and coefficient errors for 200\% added noise  are given in Table \ref{tableResults200}.
In all cases, the raw equations had the ``true'' functionals. 
Model 2 had the best train-val trajectories, the lowest  errors, and also the best test trajectories (shown in \ref{figLorenzEvolvedTestIncreasingNoise}, right column).  
Despite apparently small differences in coefficients (relative to Model 2), Model 0 had poor train-val trajectories, highlighting how variations in quantitative error do not necessarily reflect changes in behavioral error.

\begin{table}[t]
  \begin{center}
    \caption{Lorenz \textbf{200}\% noise:
Discovered  (raw) and equivalent models, and absolute coefficient errors for each true functional, $|( \hat{\xi} - \xi )  /  \xi |$ as percentage (``\textit{inf}'' indicates a missed true functional). 
``Raw'' errors are for the discovered equation, ``closest'' errors are for the transformed equation (cf section \ref{sectionLinearSpanAssessment}).
}
\label{tableResults200}
    \begin{tabular}{c|c|c|l|l} 
      \hline
     \rule{0pt}{2.0ex}   ODE   & Raw Eqn   &  Closest Eqn &  Raw Err \%  & Closest Err \%  \\  
 \hline
 Model 0& & & & \\
      $\dot{x}$  & $-5.28 x  + 5.85 y$       &    same    & (47, 41) &  (47, 41)  \\
      $\dot{y}$ & $16.04 x  + 4.67 y  -0.93 xz$ & $~24.21 x  + 2.88 y   -1.15 xz$ & (43, 567, 7)  & (14, 388, 15)  \\
      $\dot{z}$ &  $-2.22 z  + 0.86 xy $  &   same  & (17, 14)   &  (17, 14)    \\ 
\hline
 Model 1 & & & & \\
      $\dot{x}$  & $ -5.58 x  + 5.42 y$            &  same   &  (44, 46) &  (44, 46)  \\
      $\dot{y}$ & $23.92 x  + 2.77 y   -1.07 xz$ & $27.03 x  + 2.19 y  -1.15 xz$ & (70, 487, 52)   &  (3, 319, 15) \\
      $\dot{z}$ & $-2.1 z  + 0.84 xy$   &  same   & (21, 16)   & (21, 16)   \\
 \hline
 Model 2 & & & & \\
      $\dot{x}$  & $ -6.52 x  + 7.04 y$            &   same  & (35, 30)&  (35, 30)  \\
      $\dot{y}$ & $8.4 x  + 3.87 y  - 0.48 xz $  & $27.63 x   -1.01 y   -0.98 xz$ & (70, 487, 52) &(1, 1, 2) \\
      $\dot{z}$ & $-2.43 z  + 0.91xy   $ &   same  & (9, 9)   &  (9, 9)     \\
\hline
    \end{tabular}
  \end{center}
\end{table}


\subsubsection{300\% added noise}
\label{sectionResultsLorenz300Noise} 

Typical discovered models and coefficient errors for 300\% added noise are given in Table \ref{tableResults300}.
In each discovered model, $\dot{x}$ contains the linearly dependent term $xz$ ($R^2 \approx 0.97$).
Substituting it out during assessment left the ``true'' functionals $x$ and $y$.
Constant terms (e.g. in model 2's $\dot{y}, \dot{z}$) could not be substituted out.
All the discovered models  contained the correct sparse libraries (Table \ref{tableResults300}), but they had degenerate trajectories consisting of simple figure-8s. 
So in terms of finding coefficients accurate enough to generate qualitatively correct Lorenz trajectories,  the toolkit hits  a failure point somewhere between 220\% and 300\% noise. 
 
\begin{table}[t]
  \begin{center}
    \caption{Lorenz \textbf{300}\% noise:
Discovered and equivalent models, and absolute coefficient errors for each true functional, $|( \hat{\xi} - \xi )  /  \xi |$ as percentage (``\textit{inf}'' indicates a missed true functional, ``$*$'' an extra incorrect functional). 
``Closest'' errors are for the transformed equation (cf section \ref{sectionLinearSpanAssessment}). 
}
\label{tableResults300}
    \begin{tabular}{c|c|c|l|}   
      \hline
    \rule{0pt}{2.0ex}    ODE   & Raw Eqn   &  Closest Eqn   &Closest Err \% \\   
 \hline
 Model 0& & &  \\
      $\dot{x}$  & $3.5 x  + 4.38 y   -0.34 xz$       &  $-7.7 x  + 6.09 y $ & (23, 39) \\ 
      $\dot{y}$ & $9.67 x  + 4.08 y   -0.65 xz$ & $~26.16 x  + 0.02 y   -1.07 xz$  & (7, 102, 7)  \\
      $\dot{z}$ &  $-15.91   -1.36 z  + 0.81 xy$  &   same   &  ($*$, 49, 19)    \\   
\hline
 Model 1& & &  \\
      $\dot{x}$  & $ 5.59 x  + 4.09 y   -0.34 xz $            & $-5.61 x  + 5.82 y $ &  (44, 42) \\ 
      $\dot{y}$ & $14.21 x  + 4.54 y   -0.81 xz $ & $25.55 x  + 2.02 y  -1.11 xz$   & (9, 302, 11)\\  
      $\dot{z}$ & $-2.01 z  + 0.76 xy$   &  same   & (25, 24)   \\  
 \hline
 Model 2& & &  \\
      $\dot{x}$  & $ 8.32 x  + 2.22 y   -0.32 xz$            &   $ -2.18 x  + 3.52 y$ & (78, 65)  \\  
      $\dot{y}$ & $-3.06  + 21.79 x  + 1.59 y   -0.89 xz  $  & $3.06  + 27.17 x  + 0.63 y   -1.03 xz$ &($*$, 3, 163, 3)\\ 
      $\dot{z}$ & $8.91   -0.58 x   -2.05 z  + 0.75 xy    $ &   same    &  ($*$, $*$, 23, 25)    \\  
\hline
    \end{tabular}
  \end{center}
\end{table}

 
\begin{figure}[t]
\centering
\centerline{
\fbox {
 \includegraphics  [width=1\linewidth ]{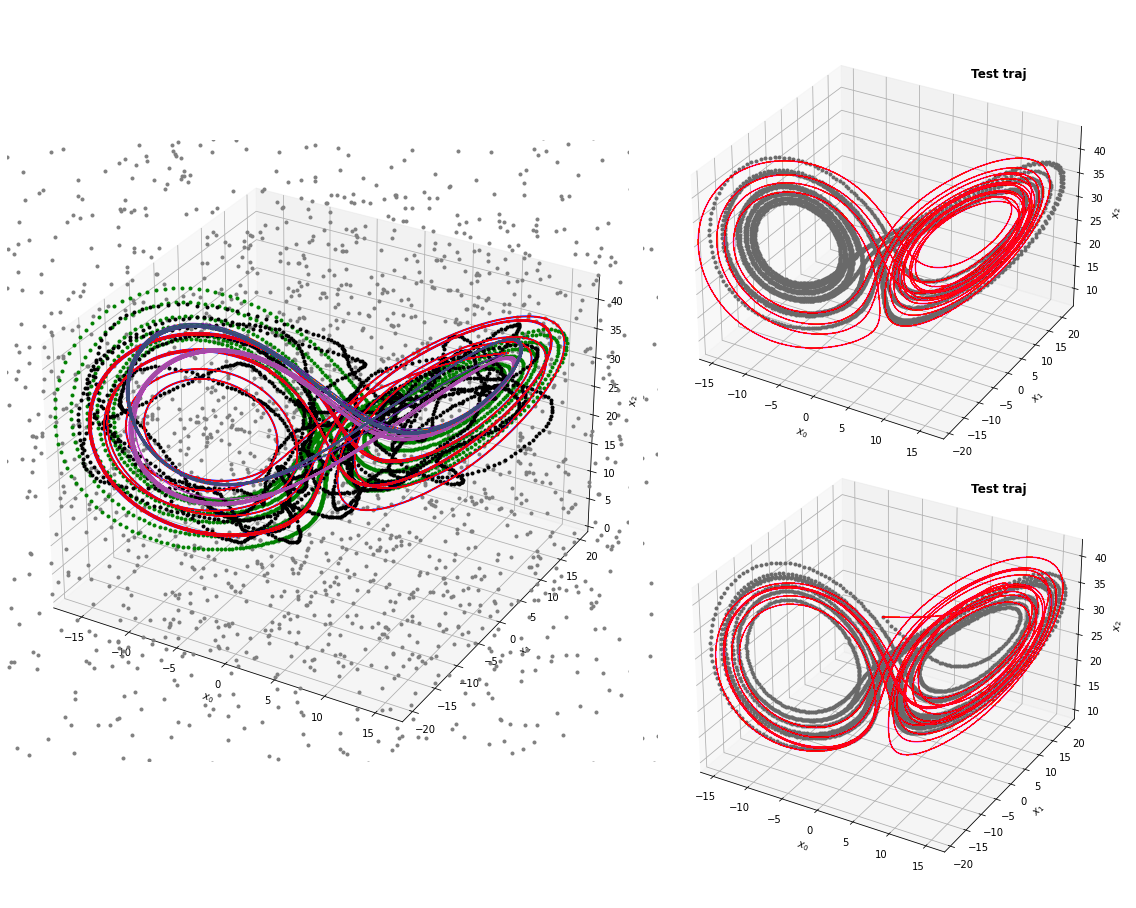} 
}
}
\caption{\small { \textbf{High noise (220 - 300\%) Lorenz, predictions of training and test trajectories.}  
{\bf{Left:}} Training trajectory:
Green = true clean trajectory; Grey dots = with 220 - 300\% added noise; Black = smoothed trajectory with artifacts.
Red = predictions (correct behavior, multiple loops per  lobe visit); 
Purples = predictions by other models (incorrect, simple figure-8s).
{\bf{Right:}} Model predictions on the two test trajectories (correct behavior). 
Grey = true trajectory, Red = predicted.
}  }
\label{figLorenzHighNoiseSmoothedTrajectories}
\end{figure}

\section{Discussion}
 
We have presented a toolkit of methods to address noisy data, for data-driven discovery of governing equations. 
While the toolkit is presented within the context of the SINDy method, many of its modules can be deployed in other, non-SINDy, architectures. 
In addition, although the various modules are strung together and presented here as a single architecture,  they are intended to be deployed separately.
Most of the modules are independent, e.g. regress on multiple sets of timepoints,; weight coefficients for culling; ignore timepoints with large functional value differences; cull via linear dependence; and use FFTs as regression targets. 
Two of the modules have dependencies:  Smoothing introduces artifacts that require mitigation by weighting timepoints and by regressing on FFTs; and restoring culled functionals requires generation of FoMs to provide triggering conditions.
 
The toolkit modules focus on (i) mitigating the effects of noise in various ways (e.g. removing timepoints from regressions; regressing over multiple sets of timepoints); and (ii) using models' predicted trajectories to assess model correctness (via FoMs and validation sets). 
The second item addresses the fact that many models can closely match a system's target derivatives while still generating poor time-series predictions.
Thus optimization and model selection based solely on matching derivatives is insufficient.

\subsection{Using linear dependence to handle non-uniqueness}
In a second main contribution, we propose a method for calculating and assessing linear dependencies between library functionals, in order to address the  inevitable non-uniqueness of models given noisy data and over-complete libraries.  This method has two goals:
First, it helps domain experts identify other valid candidate functionals beyond those found in the sparse discovered equations.
Second, it enables identification of equivalent transformations of a discovered model, in order to accurately assess whether a discovered model that appears to be wrong has in fact an equivalent form that closely matches the ``true'' model
This is important  when assessing any data-driven discovery method. 

\subsection{Role of domain expertise}
While this toolkit can be run ``plug and play'', we view it (and data-driven discovery methods in general) primarily as an aid to domain experts, to allow them to identify promising sets of discovered governing equations. 
Domain expertise plays a central role in SINDy (and other) methods, for example to choose coordinate systems and initial functional libraries \cite{Daniels2015naturecomm, Brunton2016pnas, Kaiser2018prsa}, although \cite{champion2019data} automates this to some extent; or to choose the best among several generated models \cite{Mangan2016ieee, Schaeffer2017pre}.
We posit that the need for domain expertise is inevitable and is not to be avoided or even minimized (except perhaps in control use-cases). 
For example, apparently automated methods of choosing a ``best model'' (e.g.AIC), while principled, necessarily assume some error function. 
This error function may be highly task-dependent, with generic error functions unsuitable for a given domain.   
Besides expert choice of coordinate systems and initial libraries, we call on domain expertise to select an optimal model from a sequence of models  based on Figures-of-Merit plots, train and validation evolutions, and properties of the active functionals in the models. 

\subsection{Limitations}
The need to evolve models at each step, in order to generate FoMs, introduces a risk is that the solvers can hang, upsetting completion of the algorithm. 
This is more likely with large libraries of high order polynomials.
One partial solution is to avoid evolutions when functionals are culled by linear dependence, since these culls tend to happen early, a time when the solvers also tend to hang.
Another partial solution is to set time limits on the solvers, to allow them to exit.
Both solutions have the drawbacks that  (i) we lose visibility into the effects (via FoMs) of culling certains functionals, and (ii) we lose the ``restore'' option (section \ref{sectionRestore}) for iterations lacking FoMs.

Another limitation is the slow runtime due to two things: (i) the serial evolutions of trajectories for FoMs; and (ii) the one-at-a-time culling method. 
The first issue can be mitigated somewhat by parallelizing training on different trajectories; by parallelizing evolutions of multiple trajectories by the same model (used to assess stability); and by skipping evolutions in the early culling iterations, when the models are too dense anyway.
The second issue can be addressed by culling more than one functional per iteration, which however coarsens the view offered by the FoMs since a deterioration from one iteration to the next cannot be ascribed to exactly one culled functional.
The runtime means that the FoM-related components of the toolkit, in their current form, would not work for certain use-cases (e.g.  applications requiring real-time model discovery).

Finally, the work here has not yet been applied to PDEs, rational functions, or control use-cases.  However, the methods presented have no limitations for being ported to these use cases. \\ \\ \\



 

\bibliography{sindyToolkitBibliography_minusNathanEntries.bib,ALLBIB.bib}

\section*{Acknowledgements}

J. N. Kutz is  acknowledges support from the National Science Foundation AI Institute in Dynamic Systems (Grant No. 2112085) and the Air Force Office of Scientific Research (FA9550-19-1-0011)
 
 
\section*{Appendix}

The Appendix has two main parts. 

First, we briefly give a list of (i) some additional techniques usable with SINDy, e.g. with the pySINDy package \cite{deSilva2020JOSS}), and (ii) some ideas that did not yield clear benefit.

Second, we give results for several other dynamical systems: 
3-D linear, linear harmonic oscillator, cubic harmonic oscillator, and the Hopf normal form (2-D). 
All are described in \cite{Brunton2016pnas}.
In general the toolkit gives good results on the 3-D linear and harmonic oscillator systems, and fair results on the Hopf system. 
 

\subsection{Additional methods}
 
\subsubsection{Methods for use with traditional SINDy}
\begin{enumerate}
\item Use different initial libraries for each variable:  
Current SINDy methods, e.g. \cite{deSilva2020JOSS}, use the same initial functional library for each variable. 
However, there is often reason to assign different libraries to each variable: (i) domain expertise might include or exclude certain functionals for certain variables; (ii) a first run of SINDy with a weak sparsity parameter might cull the libraries (differently for each variable), preparing for a second run with a tighter sparsity parameter. 
\item Incrementally cull functional libraries via iterative applications of SINDy:
SINDy can be run iteratively,  with progressively tighter sparsity parameters, culling a few functionals each time. 
The value of this approach is that each iteration does a new regression on the derivatives with a smaller, higher-probability library. Nuisance functionals can be rejected early before the definitive regression (with high sparsity parameter) is run.
This method appeared to improve SINDy models given some noise, though at lower levels than tolerated by the toolkit described here.
\item Smooth SINDy's derivative estimates then feed them back into SINDy:
Because noisy (especially non-continuous) derivative estimates cause such trouble, the following method tends to improve SINDy predictions by improving the derivative estimates: (i) Smooth the initial derivative estimates; (ii) fit a SINDy model; (iii) extract the derivatives generated by the sparse SINDy model; (iv) smooth them; (v) refit the SINDy model, feeding in the smoothed derivatives as the $\bm{\dot{x}}$ argin.
A possible reason this works is: the sparsifying effect of SINDy can remove noise from derivative estimates by simplifying their functional form; thus the SINDy model's derivative estimates can be cleaner than the original estimates based on the noisy data. 
\end{enumerate}


\subsubsection{Methods that had doubtful or no benefits}
\begin{enumerate}
\item Clipping, splining or shrinking noisy data points:
These methods of reducing noise were not effective, mainly (we believe) because they did not return continuous derivative estimates. 
A low-pass filter performed better.
\item Culling functionals based on the (un)reliability of their coefficient estimates:
Regressing several times over different sets of timepoints yields a distribution of coefficient estimates for each functional (cf section \ref{sectionBoosting}). 
The reliability of the estimates can be measured by, for example, $\sigma / \mu$ (std dev/mean).
Suppose that true functionals might have consistent estimates (since they have a true role) while spurious functionals  have noisy coefficient estimates (since they are spurious):
Then functionals with unreliable estimates can be preferentially culled.
However, experiments indicate that the coefficients of true and spurious functionals are entwined. 
If one functional's coefficient values swing wildly, it can cause swings in other functionals' coefficients. 
\item Regress on the outcomes of short trajectories rather than on derivatives: 
Instead of regressing on derivatives, regression can be on short predictions, e.g. over 100 timesteps. 
This tends to reward models that have a longer time horizon than models regressing on, for example, Runge-Kutta derivative estimates.
In our experiments, sometimes this alternative regression target worked better and sometimes not, with no clear pattern. 
\item Ridge or Lasso regression (vs ordinary $L_2$) did not improve results, and often gave worse results.
Lasso introduces confusion because it also enforces sparsity.
\end{enumerate}
 

\subsection{Results for other systems}

\subsubsection{Linear 3-dimensional system} 
\label{sectionResultsLinear3Dim}

\begin{figure}[t]
\centering 
\centerline{
\fbox {
\includegraphics[width=1\linewidth]{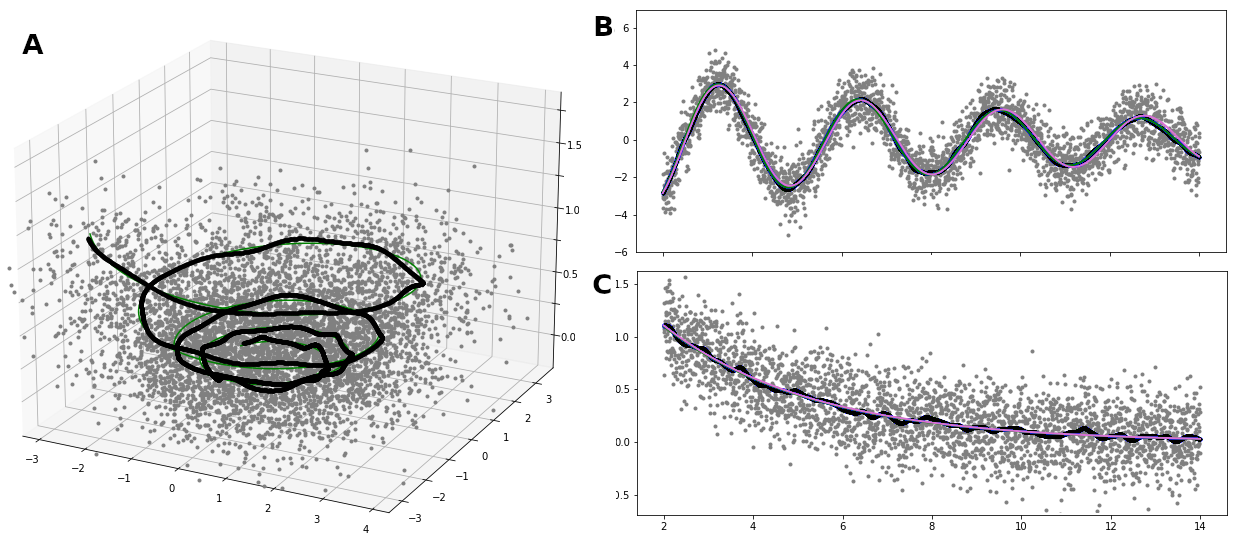}  
}
}
\caption{\small { \textbf{Linear 3-D system val trajectories.} 
{\bf{A:}} Training trajectory with 50\% noise (grey dots), clean (green), and smoothed (black).
{\bf{B:}} Time-series of $x$ ($y$ is similar but with phase shift). 
{\bf{C:}} Time-series of $z$. 
Black lines are smoothed trajectories. 
Grey dots are trajectories with added noise.
Fuschia lines are typical  predicted  validation trajectories.  
} }
\label{figLinear3D_50Noise_lib2And3_trainVal}
\end{figure}

We consider a three-dimensional linear system, with added white noise equivalent to 50\% (see Figure \ref{figLinear3D_50Noise_lib2And3_trainVal}).
Runs used libraries of either $\leq$ 2\textsuperscript{nd} or $\leq$ 3\textsuperscript{rd} degree polynomials.
Results for typical runs are reported.

The true system has ODEs:
\begin{flalign}
&\dot{x} = -0.1x - 2y \\
&\dot{y} = 2x - 0.1y\\
&\dot{z} = -0.3z
\end{flalign}
We used three training trajectories with initial conditions [2, 0, 1], [4, -1, 2], and [3, 3, 3]; and two holdout trajectories with initial conditions [3, 1, 1] and [9, 1, 3].
The initial conditions of the first training trajectory are from \cite{Brunton2016pnas}; all others were selected at random.
Trajectories were 24 seconds long with 0.002 second timestep. 

\begin{figure}[t]
\centering 
\centerline{
\fbox {
\includegraphics[width=1\linewidth]{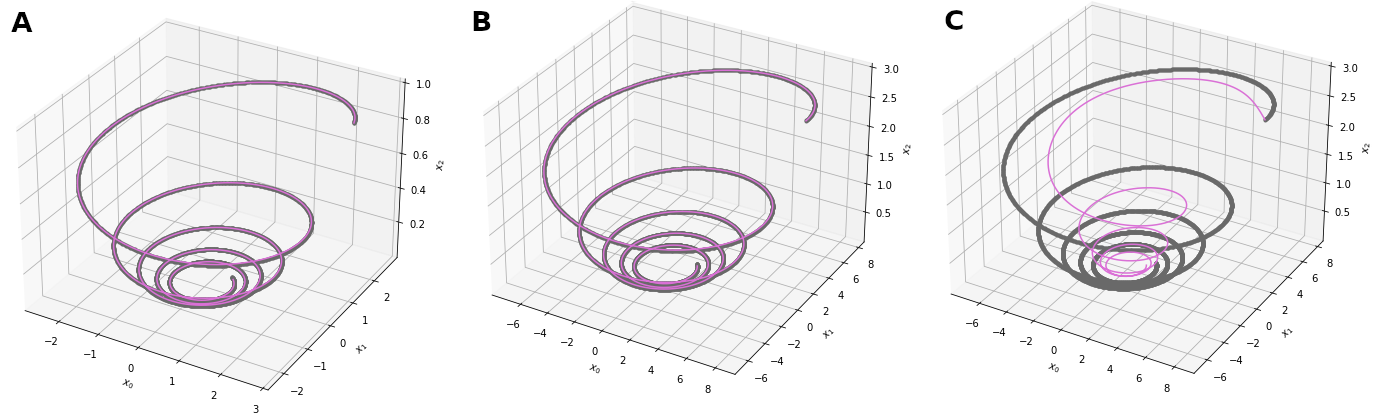}  
}
}
\caption{\small { \textbf{Linear 3-D system test trajectories.} 
{\bf{A, B:}} Typical  predictions for the two Test trajectories by models trained on data with 50\% noise  (2\textsuperscript{nd} order polynomial library).
Grey lines are true, fuchsia lines are predictions. 
The two trajectories have different initial conditions (evident in the axes scales).
{\bf{C:}} Prediction of Test trajectory 1 (as in B) by a model trained from a 3\textsuperscript{rd} order polynomial library, showing some degeneration. 
Test trajectory 0 predictions were highly accurate (similar to A).
} }
\label{figLinear3D_50Noise_lib2And3_test}
\end{figure}

At 50\% noise, discovered models were very accurate (correct libraries; median coefficient errors 1 to 8\%; accurate predicted trajectories) when the initial library contained degree 2 polynomials (Table \ref{tableResultsLinear3Dim} columns 1 and 2). 
When the initial library included degree 3 polynomials, cubic terms displaced the true minor terms in $\dot{x}$, while $\dot{y}$ and $\dot{z}$ estimates remained accurate (Table \ref{tableResultsLinear3Dim} columns 3 and 4).  
In all cases, the discovered models made highly accurate predictions for train and val  trajectories (see Fig \ref{figLinear3D_50Noise_lib2And3_trainVal}). 
Predictions of one test trajectory deteriorated somewhat given the 3\textsuperscript{rd} degree library models (see Fig \ref{figLinear3D_50Noise_lib2And3_test}C). 

At 70\% added noise (initial library $\leq 3^{rd}$ degree polynomials) the following changes occurred: 
(ii) the displacement of minor functionals increased; (iii) train and validation trajectory predictions remained accurate; (iv) test trajectory predictions deteriorated. 
The discovered major functionals of $\dot{x}$ and $\dot{y}$, as well as all of $\dot{z}$, remained accurate.
 
\begin{table}[t]
  \begin{center}
    \caption{Linear 3-dimensional model, 50\% noise:
Discovered models and coefficient errors for runs using an initial library of polynomials up to degree 2 and up to degree 3.  
Columns 2 and 4 gives the raw discovered equations for each library.
Column 3 and 5 give the absolute coefficient errors for each true functional, $|( \hat{\xi} - \xi )  /  \xi |$ as percentage (``\textit{inf}'' indicates a missed true functional , ``$*$'' an extra incorrect functional). 
}
\label{tableResultsLinear3Dim}
    \begin{tabular}{c|c|l|c|l} 
      \hline
 \rule{0pt}{2.0ex}      ODE   & using 2\textsuperscript{nd} degree library &  Error  \% &   using 3\textsuperscript{rd} degree library  &  Error \%  \\  
 \hline
 Model 0: & & & & \\
      $\dot{x}$  & $ -0.12 x   \!- \!1.94 y   \!- \!0.62 z   \!+ \! 1.66 z^2 $ &  (20, 3,  \!$*$,  \!$* \! \!$ )    & $-1.98 y  \! + \! 0.71 xz  \!  - \!0.22 x^3 $ &  (\textit{inf}, 1,  \!$*$,  \!$* \! \!$ )  \\
      $\dot{y}$ & $ 1.97 x   -0.09 y $ & (1, 6)  & $ 1.96 x    -0.13 y  $   &  (2, 28)   \\
      $\dot{z}$ & $   -0.27 z   $ & (10)    & $ -0.28 z $   &  (7)  \\
\hline
 Model 1:  & & & & \\
      $\dot{x}$  & $-0.11 x   -1.97 y  $ &  (8, 1)  &  $ -2.0 y   -0.17 xz $ &   (\textit{inf}, 0, $*$)  \\
      $\dot{y}$ & $ 2.0 x  + -0.11 y$ &  (0, 12)  & $  2.0 x   -0.11 y  $   &  (0, 8) \\ 
      $\dot{z}$ & $ -0.28 z $ & (8)  & $  -0.29 z  $   &  (2)   \\
 \hline
 Model 2:  & & & & \\
      $\dot{x}$  & $ -0.08 x  -1.99 y  $  &   (21, 1)    & $  -1.99 y    -0.02 x^3  $ & (\textit{inf}, 0, $*$)  \\
      $\dot{y}$ & $ 1.99 x   -0.12 y  $  &  (0, 24)  & $ 1.98 x   -0.12 y $&(1, 19)\\ 
      $\dot{z}$ & $  -0.3 z  $ & (1)   & $  -0.3 z     $   &  (1)   \\
\hline
    \end{tabular}
  \end{center}
\end{table}
%

\subsubsection{Linear harmonic oscillator} 
\label{sectionResultsHarmOscLinear}

\begin{figure}[t]
\centering 
\centerline{
\fbox {
\includegraphics[width=1.0\linewidth]{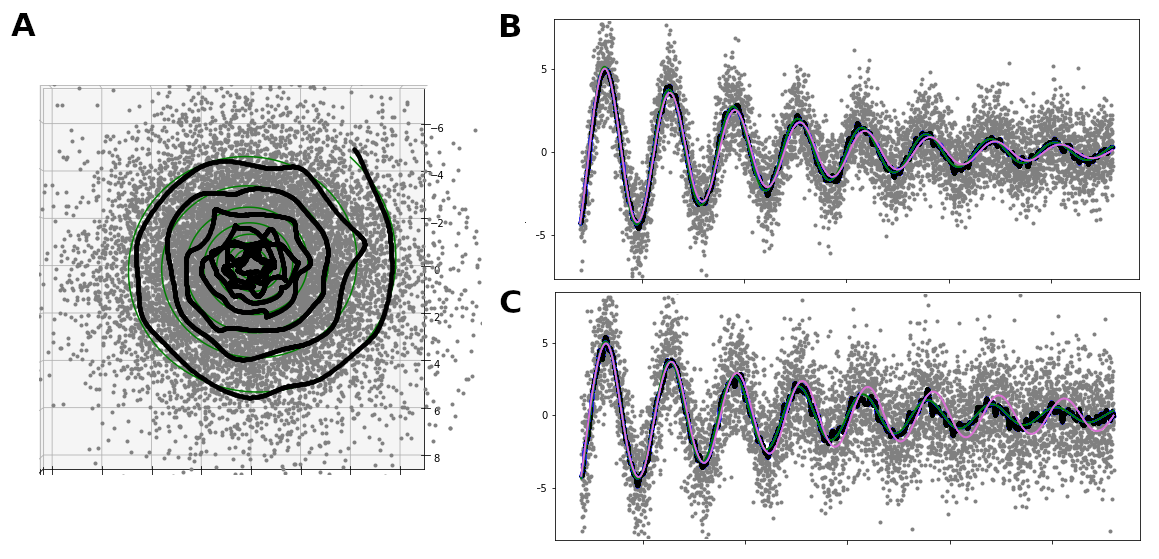}  
}
}
\caption{Harmonic linear oscillator, 70\% and 100\% noise. 
{\bf{A:}} Time-series in $x$-$y$ plane, with 100\% noise (grey dots), clean (green), and smoothed with squiggle artifacts (black).
{\bf{B:}} Typical  validation trajectory ($x$ shown, $y$ is similar but with phase shift) given 70\% noise during training.
{\bf{C:}} The same, but given 100\% noise during training.
Grey dots are trajectories with added noise.
Black lines are smoothed trajectories. 
Fuschia lines are the trajectories as predicted by discovered models.  
}
\label{figIncreasingNoiseHarmOscLinear}
\end{figure}

We consider a two-dimensional harmonic linear oscillator with added white noise equivalent to 70\% and100\% (see Figure \ref{figIncreasingNoiseHarmOscLinear}).
Results for typical runs are reported.

The true system has ODEs:
\begin{flalign}
&\dot{x} = -0.1x + 2y   \\
&\dot{y} = -2x - 0.1 y  
\end{flalign}

Three training trajectories had initial conditions [2, 0], [4, 1], and [7, 1]; and two holdout trajectories had initial conditions [3, 2] and [6, 3].
The initial conditions of the first training trajectory were from \cite{Brunton2016pnas}; all others were selected at random.
Trajectories were 28 seconds long with 0.002 second timestep.
The initial functional libraries were all polynomials with degree $\leq$ 3.
Linear dependencies between terms were weak, so the discovered models had no transformed versions. 

At 70\% noise, the method typically recovered the correct sparse functional libraries and accurate coefficients (Table \ref{tableResultsHarmOscLinear}), and gave accurate predictions for all trajectories (train, val, and test). 

At 100\% noise, the discovered models lost the $x$ term in  $\dot{x}$, keeping instead $xy^2$. 
All other functionals were correct, with accurate coefficients (Tabse \ref{tableResultsHarmOscLinear}), and  predicted trajectories (train, val, and test) were also accurate.

\begin{table}[t]
  \begin{center}
    \caption{Harmonic linear oscillator, 70 to 100\% noise:
Discovered models and coefficient errors for 70\% and 100\% noise.  
Columns 2 and 4 gives the raw discovered equations for 70\% noise and 100\% noise.
Column 3 and 5 give the absolute coefficient errors for each true functional, $|( \hat{\xi} - \xi )  /  \xi |$ as percentage (``\textit{inf}'' indicates a missed true functional , ``$*$'' an extra incorrect functional). 
At 70\% noise, Model 1 (which had $y^3$ instead of $y$) had clearly inferior validation trajectories, allowing easy identification of Models 0 and 2 as correct. 
}
\label{tableResultsHarmOscLinear}
    \begin{tabular}{c|c|c|l|l} 
      \hline
     \rule{0pt}{2.0ex}   ODE   & Raw Eqn, 70\% noise  &  Error  \% &  Raw Eqn, 100\% noise  &  Error \%  \\  
 \hline
 Model 0: & & & & \\
      $\dot{x}$  & $-0.08 x  + 2.0 y $       & (16, 0)   & $ 2.0 y   -0.18 xy^2$ & (\textit{inf}, 0) \\
      $\dot{y}$ & $-1.95 x   -0.12 y $ & (2, 17) & $-2.0 x  -0.09 y$   & (0, 11)  \\
\hline
 Model 1: & & & & \\
      $\dot{x}$  & $ -0.13 x  + 1.98 y$            & (33, 1) &  $ 1.95 y $&  (\textit{inf} 3) \\
      $\dot{y}$ & $-1.99 x  + 0.01 y^3 $ & (0, \textit{inf}, $*$) & $-1.95 x   -0.11 y$   & (3, 7)\\ 
 \hline
 Model 2: & & & & \\
      $\dot{x}$  & $ -0.11 x  + 1.94 y $            &  (11, 3)   & $  1.98 y   -0.02 xy^2$ & (\textit{inf}, 1)  \\
      $\dot{y}$ & $-2.01 x   -0.09 y  $  & (0, 9)  & $-1.95 x   -0.1 y$&(2, 0)\\ 
\hline
    \end{tabular}
  \end{center}
\end{table}

$~$\\ \\ \\

\begin{figure}[]
\centering 
\centerline{
\fbox {
\includegraphics[width=1\linewidth]{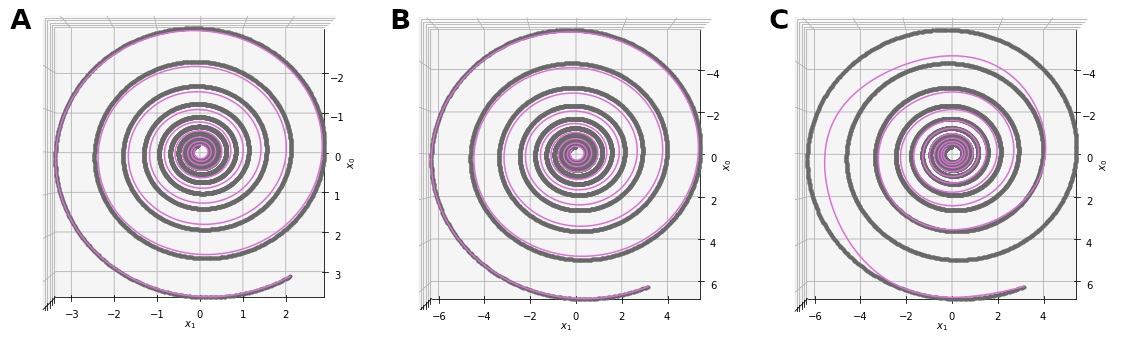}  
}
}
\caption{Harmonic linear oscillator, test trajectories.  
{\bf{A, B:}} Typical  predictions for Test trajectories 0 and 1, training  data with 70\% noise.
Grey lines are true, fuchsia lines are predictions.  
{\bf{C:}} Prediction of Test trajectory 1 (as in B), training data with 100\% noise, showing some degeneration. 
Test trajectory 0 predictions were highly accurate given training data with 100\% noise (very similar to A).
}
\label{figHarmOscLinearTestTraj}
\end{figure}


\subsubsection{Cubic harmonic oscillator}
\label{sectionResultsHarmOscCubic}

\begin{figure}[t]
\centering 
\centerline{
\fbox {
\includegraphics[width=1.0\linewidth]{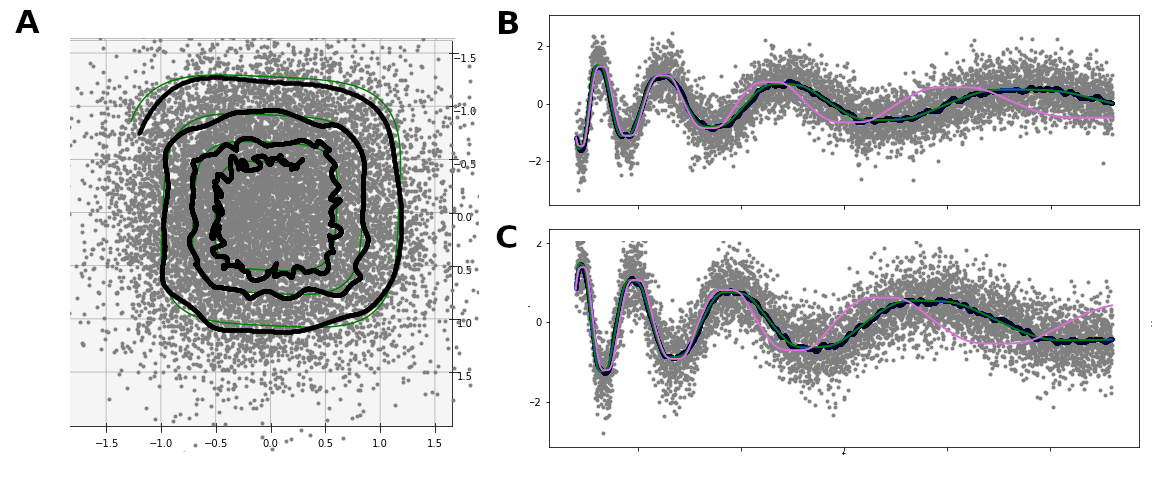}  
}
}
\caption{Harmonic cubic oscillator, 70\% noise. 
{\bf{A:}} Time-series in $x$-$y$ plane, noisy (grey dots), clean (green) and smoothed with squiggle artifacts (black).
{\bf{B:}} Typical $x$ time-series, with prediction as training trajectory.
{\bf{C:}} $x$ time-series, with prediction as validation trajectory.
Grey dots are trajectories with added noise.
Black lines are smoothed trajectories. 
Fuschia lines are the trajectories as predicted by discovered models.  
}
\label{figHarmOscCubic70TrainValTraj}
\end{figure}

We consider a two-dimensional harmonic cubic oscillator with added white noise equivalent to 70\% (see Figure \ref{figHarmOscCubic70TrainValTraj}).
Results for typical runs are reported.

The true system has ODEs:
\begin{flalign}
&\dot{x} = -0.1 x^3 + 2 y^3   \\
&\dot{y} = -2 x^3 -0.1 y^3 
\end{flalign}

We used three training trajectories with initial conditions [2, 0], [4, 1], and [7, 1]; and two holdout trajectories with initial conditions [3, 2] and [6, 3].
The initial conditions of the first training trajectory are from \cite{Brunton2016pnas}; all others were selected at random.
Trajectories were 28 seconds long with 0.002 second timestep.
The initial functional library was all polynomials with degree $\leq$ 5.

\begin{figure}[]
\centering 
\centerline{
\fbox {
\includegraphics[width=1\linewidth]{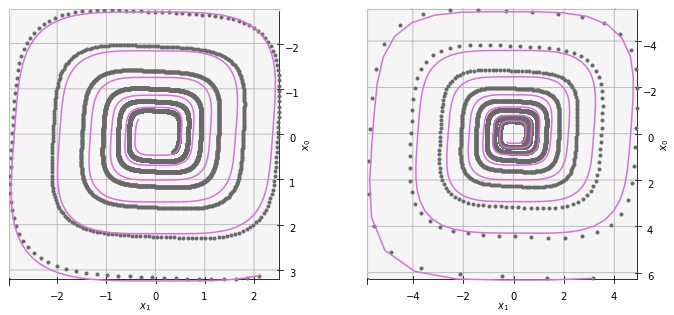}  
}
}
\caption{Harmonic cubic oscillator Test trajectories. 
Predicted test trajectories using a typical discovered model trained on data with 70\% noise. 
Grey dots are test trajectories (subsampled).
Fuschia lines are the predicted trajectories.
}
\label{figHarmOscCubic70Test3D}
\end{figure}

At 70\% noise, the method typically recovered the correct sparse functional libraries and accurate coefficients for $\dot{y}$, but missed the minority term $x^3$ in $\dot{x}$, keeping instead various 5\textsuperscript{th} order terms which had strong linear dependencies ($R^2 \approx$ 0.88 to 0.92).
See Table \ref{tableHarmCubOsc70}.
The train, val, and holdout trajectories were largely accurate (see Figs \ref{figHarmOscCubic70TrainValTraj} and \ref{figHarmOscCubic70Test3D}). 

At 100\% noise,  some models yielded correct functional libraries but results were overall unreliable, as many models included 5\textsuperscript{th} order terms. 

\begin{table}[t]
  \begin{center}
    \caption{Harmonic cubic oscillator, 70\% noise:
Discovered and equivalent models, and absolute coefficient errors for each true functional, $|( \hat{\xi} - \xi )  /  \xi |$ as percentage (``\textit{inf}'' indicates a missed true functional , ``$*$'' an extra incorrect functional). 
``Raw'' errors are for the discovered equation, ``closest'' errors are for the transformed equation (cf section \ref{sectionLinearSpanAssessment}). 
}
\label{tableHarmCubOsc70}
    \begin{tabular}{c|c|c|l|l} 
      \hline
   \rule{0pt}{2.0ex}     ODE   & Raw Eqn   &  Closest Eqn &  Raw Err \%  & Closest Err \%  \\  
 \hline
 Model 0& & & & \\
      $\dot{x}$  & $  -0.24 xy^2  + 2.04 y^3  $       &  same & (\textit{inf}, 2, $*$)  & (\textit{inf}, 2, $*$) \\
      $\dot{y}$ & $-2.03 x^3   -0.12 y^3$ & same & ( 6, 20)   & (6, 20 )  \\ 
\hline
 Model 1& & & & \\
      $\dot{x}$  & $ -0.28 xy^2 + 2.18 y^3 $            & same  &  (\textit{inf}, 9, $*$) &  (\textit{inf}, 9, $*$)  \\
      $\dot{y}$ & $-1.95 x^3   -0.15 y^3$ & same  &(3, 49)   & (3, 49)\\ 
 \hline
 Model 2& & & & \\
      $\dot{x}$  & $  1.75 y^3  + 0.33 x^4y  + 0.17 y^5 $ &   $ -0.01 x^3  + 1.98 y^3  + 0.33 x^4y $  & (\textit{inf}, 12, $*$, $*$)& (96, 1, $*$)  \\
      $\dot{y}$ & $-2.01 x^3   -0.07 y^3 $  & same & (1, 26)&(1, 26)\\ 
    \end{tabular}
  \end{center}
\end{table}
 

\subsubsection{Hopf Normal 2D}
\label{sectionHopf2D}

We consider a two-dimensional Hopf Normal form  (using the identity $z = x^2 + y^2$), as described in \cite{Brunton2016pnas} with added white noise equivalent to 70\%.
Results for typical runs are reported.

The true system has ODEs:
\begin{flalign}
&\dot{x} = 0.2 x + y  - x(x^2 + y^2)  ~~ = ~~0.2x +  y -  x^3 - xy^2 \\
&\dot{y} = x + 0.2 y - y(x^2 + y^2) ~~  = ~~x + 0.2y -  x^2y - y^3
\end{flalign}

We used three training trajectories with initial conditions [1, 0.75], [0.9, -0.1], and [0.25, 1]; and two holdout trajectories with initial conditions [0.1, -0.75], [0.5, -0.5].
The initial conditions of the first training trajectory are from \cite{Brunton2016pnas}; all others were selected at random.
Trajectories were 16 seconds long with 0.002 second timestep.
The initial functional library was all polynomials with degree $\leq$ 5.
The discovered models have two salient features.  

First, the toolkit discovered apparently incorrect libraries that were equivalent (via strong linear dependencies) to highly accurate libraries.
Examples of relevant linear dependencies are shown in  Fig \ref{figHopf2DLinearDeps}.
The equivalent models had highly similar evolved trajectory behavior, correct functional libraries,  and often very accurate coefficient estimates (median 8\% error for $\dot{y}$, though median 78\% error for $\dot{x}$).
Coefficient errors for discovered and closest (via linear dependence) models are given in Table \ref{tableHopf2DModelsErrors70}.
The equivalency of an apparently almost entirely wrong discovered model and a linear transformation very close to the ``true'' model highlights the importance of the linear dependence assessment method (\ref{sectionLinearSpanAssessment}).

\begin{figure}[t]
\centering 
\centerline{
\fbox {
\includegraphics[width=1.0\linewidth]{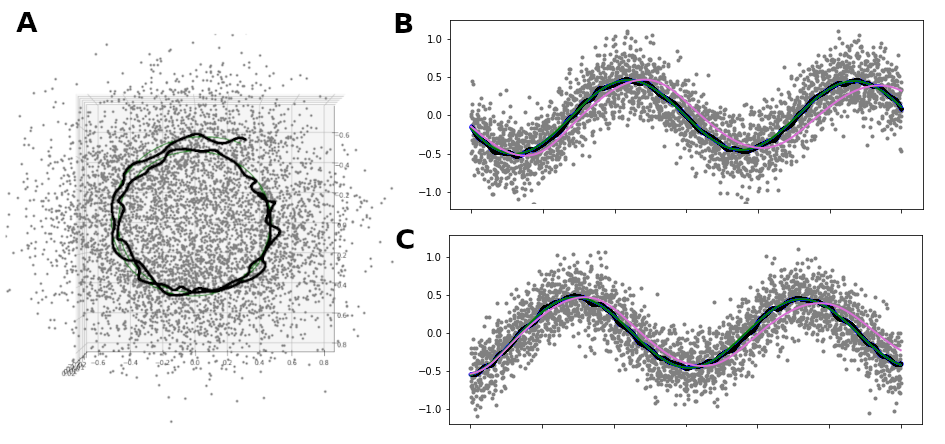}  
}
}
\caption{Hopf Normal form, 2-D,  70\% noise, training set. 
{\bf{A: }}Time-series in $x$-$y$ plane, noisy (grey dots), clean (green) and smoothed with squiggle artifacts (black).
{\bf{B, C:}} Typical $x$ time-series, with predictions for two validation trajectories.
Grey dots are trajectories with added noise.
Black lines are smoothed trajectories. 
Fuschia lines are the trajectories as predicted by discovered models.  
}
\label{figHopf2DtrainValTraj}
\end{figure}

\begin{figure}[t]
\centering 
\centerline{
\fbox {
\includegraphics[width=1\linewidth]{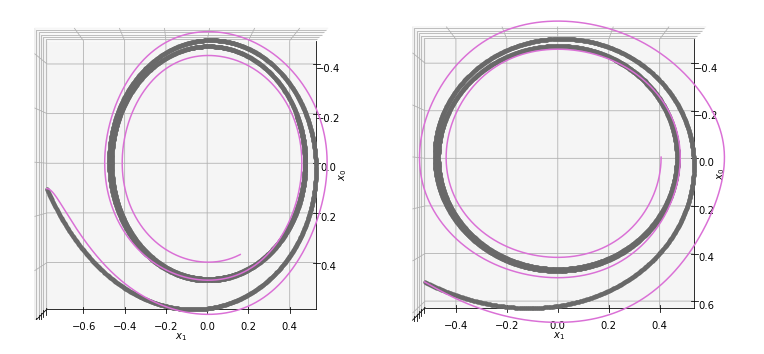}  
}
}
\caption{Hopf Normal form, 2-D, test set predictions. 
Predicted test trajectories using a typical discovered model trained on data with 70\% noise. 
Grey dots are test trajectories.
Fuschia lines are the predicted trajectories.
The trajectories are qualitatively correct, but suffer from a phase offset.
}
\label{figHopf2DTestTraj}
\end{figure}

Second, the trajectories evolved by the best models (as judged on validation trajectory FoMs and evolutions) matched the true trajectories (train, validation, and test) well though not perfectly (see Figs \ref{figHopf2DtrainValTraj}, \ref{figHopf2DTestTraj}).
This disconnect between correct equation form and correct predictive ability highlights the distinction between the goals of inference   (identifying correct functional libraries), and  prediction (producing models that behave correctly). \\ \\ \\
%
%

\begin{table}[t]
    \caption{Hopf Normal form, 2-D version, 70\% noise:
Discovered models and their coefficient errors are given in the top half of the table; closest equivalent models (cf section \ref{sectionLinearSpanAssessment}) and their errors are given in the bottom half.  
The table gives absolute coefficient errors for each true functional, $|( \hat{\xi} - \xi )  /  \xi |$ as percentage (``\textit{inf}'' indicates a missed true functional , ``$*$'' an extra incorrect functional). 
``Raw'' errors are for the discovered equations, ``closest'' errors are for the transformed equations.  
}
\label{tableHopf2DModelsErrors70}
    \begin{tabular}{c|c|l} 
      \hline
  \rule{0pt}{2.0ex}  ODE   & Raw Eqn   &   Raw Err \%    \\  
 \hline
(True:) & & \\
   $\dot{x}$  & $ 0.2x +  y -  x^3 - xy^2   $   & 0   \\
      $\dot{y}$ & $x + 0.2y -  x^2y - y^3$ & 0   \\ 
\hline
 Model 0& &   \\
      $\dot{x}$  & $ -0.9 y   -1.18 xy^2   -1.16 y^3  + 4.66 y^5  $   & (\textit{inf}, 190, \textit{inf}, 18, $*$, $*$)    \\
      $\dot{y}$ & $0.96 x$ & (4, \textit{inf}, \textit{inf}, \textit{inf})    \\ 
\hline
 Model 1& &  \\
      $\dot{x}$  & $ -1.31 y  + 1.9 y^3$             &  (\textit{inf}, 231, \textit{inf}, \textit{inf}, $*$)   \\
      $\dot{y}$ & $1.06 x  + -0.06 y   -0.54 x^3$  &(6, 131, \textit{inf}, \textit{inf}, $*$)    \\ 
 \hline
 Model 2& &  \\
      $\dot{x}$  & $ -1.34 y  -0.22 xy^2  + 2.18 y^3 $   & (\textit{inf}, 234, \textit{inf}, 78, $*$)  \\
      $\dot{y}$ & $0.98 x    - 0.27 y^3  - 0.44 x^3 $    & (2, \textit{inf}, \textit{inf}, 73, $*$) \\ 
   \hline
   \rule{0pt}{2.0ex}     ODE   & Closest Eqn & Closest Err \%  \\  
 \hline
Model 0& &   \\
      $\dot{x}$     &   $ 0.05 x  + -0.9 y  -0.21 x^3  -1.42 xy^2   -1.16 y^3  + 4.66 y^5 $ &   (75, 190,  79,  42, $*$, $*$) \\
      $\dot{y}$ &  $0.96 x  + 0.21 y   -0.88 x^2y   -0.97 y^3 $ & (4,  4,  12,   3)  \\ 
\hline
 Model 1& &  \\
      $\dot{x}$  &  $0.05 x -1.32 y  -0.22 x^3   -0.23 xy^2  + 1.9 y^3 $ &  (75,  232,   78,   77, $*$)  \\
      $\dot{y}$ &  $1.06 x  + 0.18 y   -0.54 x^3  -1.11 x^2y   -1.04 y^3$   & (6,  10,  11,   4)\\ 
 \hline
 Model 2& &  \\
      $\dot{x}$  &  $  0.05 x  -1.34 y  -0.21 x^3  -0.47 xy^2  + 2.18 y^3  $  &  (75,  234,  79,  53, $*$)  \\
      $\dot{y}$ &   $ 0.98 x  + 0.18 y    -0.77 x^2y   -1.13 y^3  -0.44 x^3 $  &(2,  10,  23,  13, $*$)\\ 
    \end{tabular}
\end{table} 
 

We note that the three-dimensional version of the Hopf Normal form (described in \cite{Brunton2016pnas}) gave the toolkit considerable trouble. 
This system includes both a transient portion and a steady-state on an attractor with $z$ = constant. 
At only 20\% added noise the models discovered by the toolkit were somewhat poor: (i) they only partially captured correct functionals, with spurious functionals and inaccurate coefficients; (ii) they had strong training trajectory predictions but poor validation trajectory predictions; and (iv) they failed to predict test trajectories.


\end{document}